%% file: main.tex
\documentclass[letterpaper, 10pt, conference]{IEEEtran}
\IEEEoverridecommandlockouts
\usepackage{float}
\usepackage{filecontents}
\usepackage{cite}
\usepackage{amsmath,amssymb,amsfonts}
\usepackage{balance}
\usepackage{algorithmic}
\usepackage{algorithm}
\usepackage{graphicx}
\usepackage[percent]{overpic}
\usepackage{textcomp}
\usepackage{soul}
\usepackage{enumerate}
\usepackage{subfigure}
\usepackage{relsize}
\usepackage{dblfloatfix}
\usepackage[dvipsnames]{xcolor}
\usepackage{tikz}
\usepackage[export]{adjustbox}

\usetikzlibrary{arrows.meta}
\usetikzlibrary{calc}
\usetikzlibrary{positioning, shapes, arrows.meta}

\newcommand{\ignore}[1]{}

\newcommand{\etal}{\textit{et al.}}

\newcommand{\n}{\hat{\mathbf{n}}}

\newcommand{\q}{\mathbf{q}}
\newcommand{\dq}{\dot{\mathbf{q}}}
\newcommand{\uu}{\hat{\mathbf{u}}}
\newcommand{\x}{\mathbf{x}}
\newcommand{\y}{\mathbf{y}}

\newcommand{\J}{\mathbf{J}}
\newcommand{\I}{\mathbf{I}}
\newcommand{\PP}{\mathbf{P}}
\newcommand{\cX}{\mathcal{X}}
\newcommand{\cY}{\mathcal{Y}}
\newcommand{\cC}{\mathcal{C}}
\newcommand{\cQ}{\mathcal{Q}}

\newcommand{\cS}{\mathcal{S}}

\newcommand{\RR}{\mathbb{R}}

\newcommand{\SEthree}{\mathrm{SE}\,(3)}

\makeatletter
\newcommand{\superimpose}[3][\mathord]{#1{\mathpalette\superimpose@{{#2}{#3}}}}
\newcommand{\superimpose@}[2]{\superimpose@@{#1}#2}
\newcommand{\superimpose@@}[3]{%
  \ooalign{%
    \hfil$\m@th#1#2$\hfil\cr
    \hfil$\m@th#1#3$\hfil\cr
  }%
}
\makeatother

\usepackage{pgfplots}
\usepackage{pgfplotstable}
\usepackage{pgfplots}
\usepgfplotslibrary{groupplots} 
\pgfplotsset{compat=1.17}
\pgfplotscreateplotcyclelist{colorblind friendly}{
    {blue, thick}, 
    {orange, thick}, 
    {cyan, thick} 
    {purple, thick}, 
    {green, thick}, 
    {red, thick}, 
    {yellow, thick}, 
}

\pgfplotscreateplotcyclelist{colorblind-friendly}{
    {blue, solid, line width=1pt},
    {orange, dashed, line width=1pt},
    {green, dashdotted, line width=1pt},
    {purple, densely dotted, line width=1pt},
    {red, densely dashed, line width=1pt},
    {brown, densely dashdotted, line width=1pt},
    {pink, loosely dashed, line width=1pt},
    {gray, loosely dashdotted, line width=1pt}
}

\def\BibTeX{{\rm B\kern-.05em{\sc i\kern-.025em b}\kern-.08em
    T\kern-.1667em\lower.7ex\hbox{E}\kern-.125emX}}
\begin{document}

\title{\LARGE \bf
ODE Methods for Computing One-Dimensional Self-Motion Manifolds
}

\author{Authors
\author{Dominic Guri$^{1}$ and George Kantor$^{1}$
\thanks{This work was supported in part by NSF Robust Intelligence 1956163, NSF/USDA-NIFA AIIRA AI Research Institute 2021-67021-35329, and USDA-NIFA LEAP 2024-51181-43291.}
\thanks{$^{1}$Dominic Guri and George Kantor are with the Robotics Institute, Carnegie Mellon University, Pittsburgh 15206 USA
    \tt\small \{dguri, kantor\}@cs.cmu.edu}}
}

\maketitle
\thispagestyle{empty}
\pagestyle{empty}

\raggedbottom
\setlength{\parskip}{0em}   
\setlength{\abovecaptionskip}{2pt}   

\setlength{\textfloatsep}{3pt} 

\setlength{\intextsep}{0pt} 

\setlength{\floatsep}{0pt} 

\maketitle

\begin{abstract}
Redundant manipulators are well understood to offer infinite joint configurations for achieving a desired end-effector pose. The multiplicity of inverse kinematics (IK) solutions allows for the simultaneous solving of auxiliary tasks like avoiding joint limits or obstacles. However, the most widely used IK solvers are numerical gradient-based iterative methods that inherently return a locally optimal solution. In this work, we explore the computation of self-motion manifolds (SMMs), which represent the set of all joint configurations that solve the inverse kinematics problem for redundant manipulators. Thus, SMMs are global IK solutions for redundant manipulators. We focus on task redundancies of dimensionality 1, introducing a novel ODE formulation for computing SMMs using standard explicit fixed-step ODE integrators. We also address the challenge of ``inducing'' redundancy in otherwise non-redundant manipulators assigned to tasks naturally described by one degree of freedom less than the non-redundant manipulator. Furthermore, recognizing that SMMs can consist of multiple disconnected components, we propose methods for searching for these separate SMM components. Our formulations and algorithms compute accurate SMM solutions without requiring additional IK refinement, and we extend our methods to prismatic joint systems --- an area not covered in current SMM literature. This manuscript presents the derivation of these methods and several examples that show how the methods work and their limitations.
\end{abstract}
 

\section{INTRODUCTION}
The value of redundant manipulators lies in their capacity to reach an end-effector target through infinite joint configurations. Hence, redundant manipulators, robot arms with more degrees of freedom than strictly necessary for the primary task, can reach an end-effector target while fulfilling auxiliary objectives such as avoiding joint limits, preventing collisions, and maximizing dexterity. However, as auxiliary tasks increase, computing a feasible joint configuration becomes more challenging, and standard inverse kinematics methods may fail even when a solution exists.

The complete set of inverse kinematics solutions for a given end-effector pose is known as a \textit{self-motion manifold} (SMM)~\cite{burdickCharacterizationControlSelfmotions1989}. SMMs are global IK solutions essential for motion planning, robot performance evaluation, and task-based design. However, in practice, SMMs are challenging to compute, leading practitioners to rely on iterative, gradient-based numerical IK solvers, which can only converge to locally optimal solutions.

In this work, we address the challenge of computing self-motion manifolds for manipulators with one degree of redundancy relative to the task. Although higher redundancy manipulators exist, commercially available redundant manipulators are typically 7DOF systems used for 6DOF tasks, and even ostensibly ``\textit{non-redundant}'' 6DOF manipulators are frequently deployed in naturally 5DOF tasks. Therefore, by improving the computation of self-motion manifolds, we contribute to a fundamental robotics concept and boost the performance of widely available robot systems.

\color{black}

\section{RELATED WORKS}
There are several important reasons for computing self-motion manifolds (SMMs), chief among them being the generation of globally optimal plans. This is particularly valuable in cluttered conditions, where numerical solvers often falter. For example, computing self-motion manifolds or at least identifying its largest usable segments ensures that the manipulators are more resilient to failure~\cite{groomRealtimeFailuretolerantControl1999,almarkhiMaximizingSizeSelfMotion2019}. SMM-based plans are also used to generate trajectories that are more resilient to joint failure~\cite{lewisFaultTolerantOperation1997,xieMaximizingProbabilityTask2022}. In the context of robot design optimization, SMM offer a robust approach to computing performance metrics that provide a comprehensive representation of design performance --- an advantage over locally optimal IK solvers. Another popular application of SMMs is in obstacle avoidance~\cite{tatliciogluAdaptiveControlRedundant2009,yaoMotionPlanningAlgorithms2010,maaroofGeneralSubtaskController2012,pingStudyKinematicsOptimization2006,maaroofPhysicalHumanRobotInteraction2016}. For example, Zhang \etal~\cite{zhangNovelDivisionBased2007} introduced a unique algorithm that involves splitting a manipulators into two parts, a distal chain that remains unchanged and a proximal chain that is altered within the task's SMM so that the whole manipulator does not collide into obstacles. However, a vast majority of obstacle avoidance works use standard numerical methods augmented with null space projection for auxiliary tasks which is computationally faster and avoids directly computing the underlying SMM~\cite{dasInverseKinematicAlgorithms1988,escandeHierarchicalQuadraticProgramming2013,havilandManipulatorDifferentialKinematics2023}.

First introduced by Burdick~\cite{burdickCharacterizationControlSelfmotions1989}, self-motion manifolds for reaching a task space configuration $\mathbf{x}$ is a configuration set $\mathcal{S}_\mathbf{x}$ defined as
\begin{align}
    \mathcal{S}_\mathbf{x} &=
    \Big\{
        \mathbf{q} \in \mathcal{C}~|~\mathbf{x} = \phi(\mathbf{q}) \in \mathcal{X},~
        \dim(\mathcal{C}) > \dim(\mathcal{X})
    \Big\} \label{eqn:smm} \\
    \mathsf{T}_\mathbf{q}\mathcal{S}_\mathbf{x} &=
    \Big\{
        \dot{\mathbf{q}} \in \mathsf{T}_\mathbf{q}\mathcal{C}~|~\mathbf{J}(\mathbf{q})\dot{\mathbf{q}} = 0
    \Big\} \label{eqn:internal-motion-space}
\end{align}
where \(\mathcal{X}\) and \(\mathcal{C}\) are the task and joint configuration spaces, respectively, and \(\phi\) is the forward kinematics function. The SMM tangent bundle, \(\mathsf{T}_\mathbf{q}\mathcal{S}_\mathbf{x}\), also known as the \textit{internal motion space}, is the set of joint velocities that maintain stationary end-effector. Such motion is sometimes called \textit{self-motion}. 

For joint configuration and task space dimensionalities \(\dim(\cC)\!=\!n\) and \(\dim(\cX)\!=\!m\), the dimensionality of the SMM \(k\!=\!\dim(\cS_\x)\!=\!n\!-\!m\), and is strictly positive because SMMs are defined only for redundant manipulators, i.e., for systems where \(n\!>\!m\).

Burdick~\cite{burdickCharacterizationControlSelfmotions1989} also hypothesized that SMMs occur as multiple disconnected component sets whose multiplicity cannot exceed the maximum number of unique IK solutions from a non-redundant manipulator. In other words, for an $n$DOF manipulator assigned to reach an $m$DOF end-effector pose (with $n\!>\!m$), the number of SMM components cannot exceed the maximum number of distinct IK solutions of an $m$DOF manipulator reaching the same end-effector pose. Therefore, a planar RRR manipulator reaching a position $(x,y)$ is limited to 2 SMM components --- since an RR manipulator can only generate two unique IK solutions. A 7DOF spatial manipulator, reaching an $\SEthree$ end-effector pose, is similarly limited to at most 16 SMM components, the theoretical limit of a 6DOF spatial manipulator.

Self-motion manifolds of dimensionality one ($k\!=\!1$) are typically computed as traces --- sets of points that parameterize curves in the joint space --- much like numerical solutions to ODEs. These traces are typically generated iteratively using the equation
\begin{align}
    \mathbf{q}_{k+1} &= \q_k + \gamma \hat{\mathbf{n}}(\mathbf{q}_k) + \mathbf{J}^\dagger(\mathbf{q}_k) \Delta\mathbf{x}_k \label{eqn:typical-ode}
\end{align}
where $\gamma$ is the integration step, $\n(\q)$ is the unit vector in the internal motion space for configuration $\q_k$, and $\J^\dagger(\q_k)$ is the pseudo-inverse of the manipulator Jacobian~\cite{zhangNovelDivisionBased2007,yaoMotionPlanningAlgorithms2010,almarkhiMaximizingSizeSelfMotion2019,xieMaximizingProbabilityTask2022}. This formulation is relatively straightforward. However, it requires careful handling of the directional ambiguity when computing $\n(\q_k)$. The update can be viewed as two parts: (a) Euler integration component, $\gamma\n(\q_k)$, for advancing along the SMM, and (b) end-effector error projection component $\J^\dagger(\q_k)\Delta\x_k$ that is intended to prevent any drift in the end-effector pose. It is worthwhile to reconsider Equation~\ref{eqn:typical-ode} and develop an improved formulation that can take advantage of advanced ODE solvers while potentially eliminating the need for additional end-effector error correction.

Other methods for computing SMMs include a grid search approach by Peidr\'o \etal~\cite{peidroMethodBasedVanishing2018}, which identifies SMM configurations by discretizing the joint space. More recently, Wu \etal~\cite{wuNovelMethodComputing2023} adopted a hypercube grid representation of the joint configuration space, combined with cellular automata rules, their method that scales to SMMs of higher order redundancies.

\subsection{Contributions}
In this work, we offer four new contributions:
\begin{enumerate}[(a)]
    \item a novel ODE formulation for computing one-dimensional self-motion manifolds (SMMs) using standard ODE solvers without the need for additional end-effector error correction,
    \item search algorithms for identifying all distinct SMM components corresponding to a specified end-effector pose,
    \item a new approach for introducing SMMs in non-redundant \(n\)DOF manipulators operating on \((n-1)\)DOF tasks (e.g., a 6DOF spatial manipulator assigned to a 5DOF task),
    \item a demonstration of our methods applied to prismatic jointed systems, as illustrated using 6DOF spatial manipulators mounted on linear rails.
\end{enumerate}
The remainder of this manuscript is structured as follows: Section~\ref{sec2:problem} introduces the problem of computing self-motion manifold and our proposed methods; Section~\ref{sec3:experiments} describes the experiments we used to validate our methods, and
Section~\ref{sec4:results} presents the results. Section~\ref{sec5:conclusion} is the conclusion.

\section{PROBLEM FORMULATION}\label{sec2:problem}
Our goal is to compute a self-motion manifold (SMM) for a manipulator with a single degree of redundancy, i.e., $k=1$. To achieve this, we formulate an ordinary differential equation (ODE) in the internal motion space of the redundant manipulator that builds Equation~\ref{eqn:jacobian-kernel-ode} which defines the unit kernel vector of the manipulator’s Jacobian:
\begin{align}
    \hat{\mathbf{n}}(\mathbf{q}) &
        = \frac{\ker(\J (\mathbf{q}))}{\Vert \ker(\mathbf{J} (\mathbf{q})) \Vert}
        \in\RR^n\label{eqn:jacobian-kernel-ode}
\end{align}

However, defining the ODE as 
\(\dot{\mathbf{q}}=\hat{\mathbf{n}}(\mathbf{q})\)
suffers from a major problem: the computation of \(\hat{\mathbf{n}}(\mathbf{q})\) is inherently directionally ambiguous, meaning its sign may flip unpredictably between successive samples along the SMM. This inconsistency renders the ODE unsolvable in practice.

To resolve this directional ambiguity, we introduce a \textit{directionally regularized ODE}:
\begin{align}
    \dot{\mathbf{q}} 
    &= g(\mathbf{q}; \hat{\mathbf{n}}_{\text{REF}})
    =
    \begin{cases}
        \hat{\mathbf{n}}(\mathbf{q}), & \text{if } \langle \hat{\mathbf{n}}(\mathbf{q}), \hat{\mathbf{n}}_{\text{REF}} \rangle > 0, \\
        -\hat{\mathbf{n}}(\mathbf{q}), & \text{otherwise,}
    \end{cases}
    \label{eqn:regularized-ode}
\end{align}
where the velocity \(\dot{\mathbf{q}}\) is chosen to be as close as possible to a given reference heading \(\hat{\mathbf{n}}_{\text{REF}}\).

Assuming a Runge-Kutta method defined by its Butcher tableau (with weights \(b_i\), nodes \(c_i\), and coefficients \(a_{ij}\)), a trace of the SMM is computed as:
\begin{align}
    \mathbf{q}_{n+1} &= \mathbf{q}_n + h \sum_{i=1}^{s} b_i\, k_i, \\
    k_i &= g\Biggl(\mathbf{q}_n + h\sum_{j=1}^{i-1} a_{ij}\, k_j;\; \hat{\mathbf{n}}_{\text{REF}} = \hat{\mathbf{n}}_n\Biggr),
    \label{eqn:rk-stage}
\end{align}
where \(s\) is the number of stages, \(h\) is the integration step, and \(k_i\) are the intermediate stage values. The reference direction \(\hat{\mathbf{n}}_n\) is updated at each step as follows:
\begin{align}
    \hat{\mathbf{n}}_n = 
    \begin{cases}
        g(\mathbf{q}_n; \hat{\mathbf{n}}_{n-1}), & \text{if } n > 0, \\
        \hat{\mathbf{n}}(\mathbf{q}_0), & \text{otherwise.}
    \end{cases}
    \label{eqn:update-ref}
\end{align}
The initial reference direction \(\hat{\mathbf{n}}_0\) is set using Equation~\ref{eqn:jacobian-kernel-ode}. For a given end-effector pose \(\mathbf{x}\), the initial configuration \(\mathbf{q}_0\) is obtained by solving an unconstrained IK problem using standard methods.

We call this regularized ODE problem a \textit{self-motion manifold initial value problem} (SMM-IVP), and we used an RK5 scheme with \(h=0.05\) for our experiments. The initial condition for the SMM-IVP is obtained by solving an unconstrained inverse kinematics (IK) problem --- ignoring joint limits, collisions, and obstacles --- and the ODE is integrated until the joint configurations return to within an integration step from the initial joint configuration. Since Equation~\ref{eqn:regularized-ode} describes an autonomous ODE, the integration step directly controls the distance advanced along the SMM per integration step, and the product of the number of integration steps and the step size is the total SMM length.

\subsection{Induced-Redundancy Self-Motion Manifolds}
There is a wide range of tasks assigned to non-redundant (i.e., 6DOF) manipulators that are inherently 5DOF. Examples include two-finger grasping and industrial applications such as polishing or drilling. Representing these tasks in a full 6DOF space is unnecessarily constraining and limits performance. We introduce a self-motion manifold formulation we call an \textit{induced redundancy SMM} that is designed to exploit such task space redundancies.

Assuming the non-redundant manipulator has forward kinematics function $\x=\phi(\q)\in\cX$, with $\dim(\cX)=n$. The task is represented by $\y=\xi(\x)\in\cY\subset\cX$, where $\dim(\cY)=n-1$. The mapping $\xi:\cX\mapsto\cY$ selects $n-1$ dimensions from $\cX$. 
A self-motion manifold for such a task is $\cS_\y$ because the invariance is in the task space $\cY$.

The function $\xi$ is an orthogonal projection onto $\cY$. There also exists a invariance (symmetry) subspace $\uu\in\cX$ , such that under a small perturbation $\delta$, \(\xi(\x + \delta\uu) = \xi(\x)\).
Therefore, the task's forward kinematics function is a composite mapping $\psi=\xi \circ \phi: \cC\mapsto\cY$, with linearization
\begin{align}
    \dot{\y} &= \J_\psi(\q) \cdot \dq, \text{ where }
    \J_\psi(\q) = \J_\xi(\phi(\q)) \cdot \J(\q).
\end{align}
Because of linearity and symmetry properties of $\xi$, its Jacobian $\J_\xi(\x)=\PP$, where $\PP$ is symmetric ($\PP=\PP^\top$) and idempotent ($\PP^2=\PP$) orthogonal projection operator with kernel $\ker(\PP)=\text{span}\{\uu\}$; hence, $\PP$ is can be defined from $\uu$ as follows
\begin{align}
    \J_\xi(\phi(\q)) &= \PP = \I - \uu\uu^\top.
\end{align}
To exploit this redundancy using the SMM-IVP formulation,
the kernel of the task Jacobian $\J_\psi$ is computed as follows
\begin{align}
    \mathbf{n}(\q;\uu)
        &= \ker(\PP \cdot \J(\q))
        = \J^\dagger(\q) \cdot \ker(\PP) \nonumber \\
    &= \J^\dagger(\q) \cdot \ker(\I-\hat{\mathbf{u}}\hat{\mathbf{u}}^\top) \\
    \n(\q;\uu) &= \frac{\mathbf{n}(\q;\uu)}{\Vert \mathbf{n}(\q;\uu) \Vert}.\label{eqn:selection-matrix}
\end{align}

Since the manipulator Jacobian maps joint velocities \(\dot{\mathbf{q}}\) to the end-effector velocity \(\mathbf{v} = (v_x, v_y, v_z, \omega_x, \omega_y, \omega_z)\), a redundancy direction \(\mathbf{u} = (0,0,0,0,1,0)\) corresponds to a symmetry with respect to rotations about the \(y\)-axis.
Additionally, $\uu$ can also be expressed in the manipulator's tool frame, if the tool frame manipulator Jacobian ${}^b\J(\q)$ and its pseudo-inverse ${}^b\J^\dagger(\q)$ are used in place of $\J(\q)$ and $\J^\dagger(\q)$, respectively.
Combined with the directionally regularized formulation (Equation~\ref{eqn:regularized-ode}), the induced redundancy mechanism generates the following directionally consistent ODE:
\begin{align}
    \dot{\mathbf{q}} 
    &= g(\mathbf{q}; \hat{\mathbf{n}}_{\text{REF}},\uu) \nonumber \\
    &=
    \Bigg\{
    \begin{array}{lr}
        \hat{\mathbf{n}}(\mathbf{q};\uu), & \text{if } \langle \hat{\mathbf{n}}(\mathbf{q};\uu), \mathbf{\hat{n}}_{\text{REF}} \rangle > 0 \\
        -\hat{\mathbf{n}}(\mathbf{q};\uu), & \text{otherwise.}
    \end{array}\label{eqn:ir-regularized-ode}
\end{align}

\begin{figure}
    \centering
    \subfigure[Single component SMM.]
        {\includegraphics[width=0.48\linewidth]{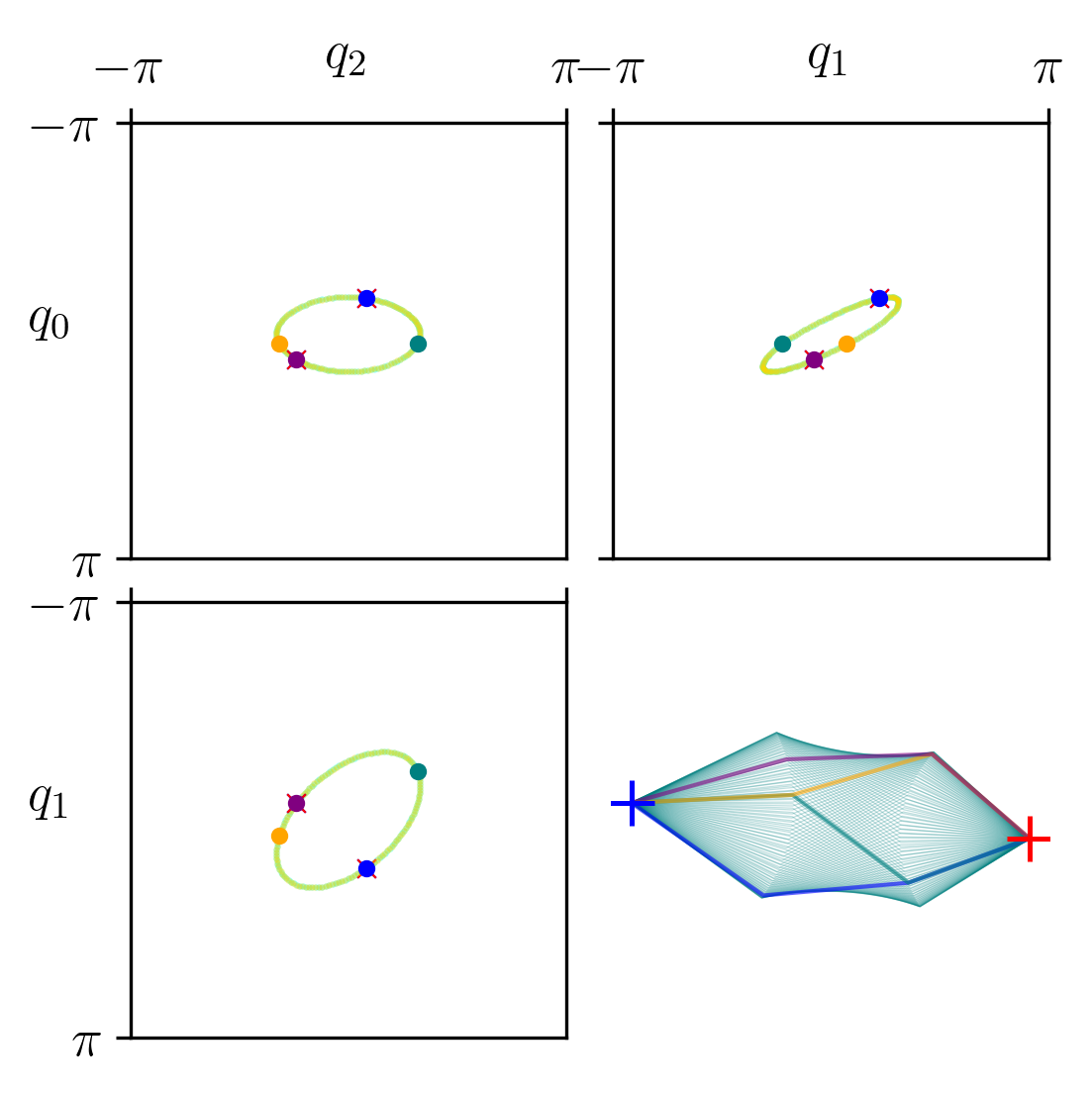}\label{fig:toggling-one-smm-a}}
    \subfigure[Two components SMM.]
        {\includegraphics[width=0.48\linewidth]{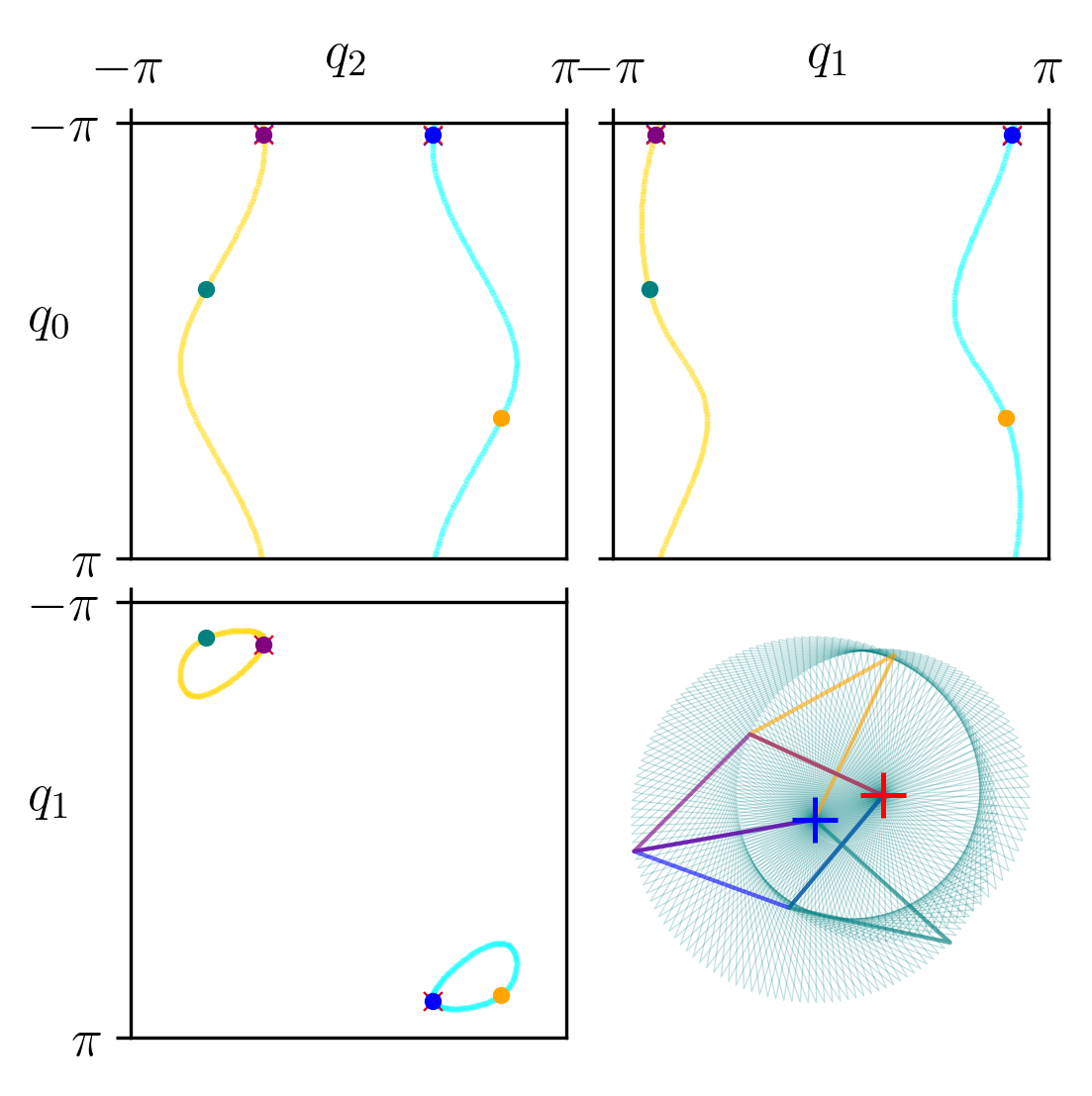}\label{fig:toggling-two-smm-a}}
        \caption{The plots above display a single-component SMM (Figure~\ref{fig:toggling-one-smm-a}) and a two-component SMM (Figure~\ref{fig:toggling-two-smm-a}) derived from \(\mathcal{Q}^*\) joint configurations (teal, blue, orange, and purple dots with matching 2D pose plots) obtained via the joint toggling process shown in Figure~\ref{fig:toggle-illutration}. In the single-component SMM, all \(\mathcal{Q}^*\) configurations lie on one trace, while in the two-component SMM they are evenly split.}\label{fig:joint-toggling-smm-demo}
\end{figure}

\subsection{Mutliple Self-Motion Manifold Components}\label{sec:QstarRRR}

\begin{algorithm}
\caption{SMM Component Search}\label{alg:smm-component-search}
\begin{algorithmic}[1]
\REQUIRE~Input $\cQ^*$, ODE $g$, RK5 integration step $h$
\ENSURE~Output result $\mathcal{S}_\mathbf{x}$
    \STATE~$\mathcal{S}_\mathbf{x} = \big\{~\mathsf{SMMIVP}(\mathbf{q}_0, g)$ \big\}
\FOR{$\q \in \cQ^*$}
    \IF{$d(\q, \cS_\x) > h$}
        \STATE~$\mathcal{S}_\mathbf{x} \gets \mathcal{S}_\mathbf{x} \cup \mathsf{SMMIVP}(\mathbf{q}_i, g)$
    \ENDIF{}
    \ENDFOR{}
\RETURN~$\mathcal{S}_\mathbf{x}$
\end{algorithmic}
\end{algorithm}

Our proposed SMM component search methods are based on Algorithm~\ref{alg:smm-component-search}. The algorithm computes SMM components using the SMM-IVP formulation, with initial values drawn from a configuration set \(\cQ^*\). Ideally, this set is the smallest collection of joint configurations from all SMM components. Sections~\ref{sec-RRR-Qstar} and~\ref{sec-Qstar-general} address the challenge of generating such a set for an RRR manipulator and for more general contexts. 

To avoid redundant computations, Algorithm~\ref{alg:smm-component-search} first checks whether a configuration is already represented in a previously computed SMM component. Specifically, a configuration \(\q\) is considered represented if the distance \(d(\q,\cS_\x)\) (see Equation~\ref{eqn:point-to-set-distance}) is less than the integration step \(h\); in practice, any random configuration from a given SMM component will be within \(\tfrac{1}{2}h\) of some configuration in the SMM-IVP trace solution.
\begin{align}
    d(\q,\cS_\x) &= \min_{\mathbf{p}\in\cS_\x} d(\q,\mathbf{p})\label{eqn:point-to-set-distance}
\end{align}
Building on Algorithm~\ref{alg:smm-component-search}, we propose two methods for generating \(\mathcal{Q}^*\): one tailored for a planar RRR robot model that efficiently guarantees all components are obtained, and another for general SMM component search.

\subsubsection{SMM Component Search for Planar RRR Robot Models}\label{sec-RRR-Qstar}
\begin{figure}[H]
    \centering
    \begin{tikzpicture}
        \begin{axis}[
                axis equal image,
            ticks=none,
            scale only axis=true,
            width=0.25\textwidth,
            axis lines=none,
            tick style={draw=none},
            every axis x label/.style={at=(current axis.south),anchor=south},
            every axis y label/.style={at=(current axis.west),anchor=west},
            xlabel={$x$},
            ylabel={$y$},
            legend pos=outer north east,
            legend image code/.code={
              \draw[#1, mark=none] (0cm,0cm) -- (0.5cm,0cm);
            },
            legend style={draw=none},
        ]
        \addplot [thick, solid, mark=none] table [x=x1, y=y1, col sep=comma, /pgf/number format/read comma as period]{poses.csv};
            \addlegendentry{$\q_0$};
        \addplot [thick, dotted, mark=none] table [x=x0, y=y0, col sep=comma, /pgf/number format/read comma as period]{poses.csv};
        \addlegendentry{$\q_1$};
        \addplot [thick, dotted, mark=none] table [x=x2, y=y2, col sep=comma, /pgf/number format/read comma as period]{poses.csv};
        \addlegendentry{$\q_2$};
        \addplot [thick, solid, mark=none, draw opacity=0.35] table [x=x3, y=y3, col sep=comma, /pgf/number format/read comma as period]{poses.csv};

        \addplot [only marks, mark=+, mark size=5pt, red, ultra thick] coordinates {(0,0)};
        \addplot [only marks, mark=+, mark size=5pt, blue, ultra thick] coordinates {(2.4675,-0.2215)};
        \addlegendentry{$\q_3$};
        \end{axis}
    \end{tikzpicture}
    \caption{This figure shows how the set $\mathcal{Q}^*$ is computed from $\q_0$ --- $\q_1$ and $\q_2$ are obtained by toggling the elbow joints, while $\q_3$, in this case, is obtained from $\q_2$. As demonstrated in Figure~\ref{fig:joint-toggling-smm-demo}, $\q_3$ need not be included for SMM component search.}\label{fig:toggle-illutration}
\end{figure}
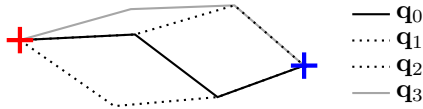

We propose a systematic approach for constructing the configuration set \(\cQ^*\) for planar robot models that guarantees capturing all SMM components. Our method begins with an initial joint configuration \(\q_0\) and then toggles the elbow joints up and down, as illustrated in Figure~\ref{fig:toggle-illutration}. In an RRR manipulator, this process initially produces four unique solutions. As demonstrated in~\cite{burdickCharacterizationControlSelfmotions1989}, if the end-effector is located within a radius of \(\ell_0+\ell_1\) (the sum of the first two link lengths), the resulting SMM will consist of a single component; otherwise, the SMM will split into two distinct components. Our approach confirms these findings, leading to the computation of the relevant SMMs (see Figures~\ref{fig:toggling-one-smm} and~\ref{fig:toggling-two-smm}). 

Based on this behavior, we determine that generating three configurations is sufficient for an exhaustive SMM component search. In our work, we therefore use \(\cQ^* = \{\q_0,\q_1,\q_2\}\), omitting \(\q_3\) because toggling the first and second elbows starting from \(\q_0\) (to obtain \(\q_1\) and \(\q_2\), respectively) already covers the necessary variations.

\subsubsection{General SMM Component Search}\label{sec-Qstar-general}

Our general approach to searching for SMM components employs Algorithm~\ref{alg:smm-component-search} as a rejection sampling mechanism. We generate the configuration set \(\mathcal{Q}^*\) by running a randomly initialized gradient-based IK solver \(N\) times, yielding \(N\) samples. No constraints are imposed during this process, ensuring that \(\mathcal{Q}^*\) broadly explores all SMM components as much as possible. In our experiments we used $N=150$.

\color{black}

\begin{table*}
    \centering
    \caption{This table presents all the details associated with the robot model used to validate the SMM-IVP solvers presented in this work. The xarm7 and RRR models are used to validate to features, while the rest are used for one.}\label{tab:experiments}
    \begin{tabular}{|l|l|c|c|c|c|c|} \hline
Test  & \textbf{Case}  & $n$\textbf{DOF}& \textbf{Task}         & \textbf{Task} $m$\textbf{DOF}&  \textbf{Features} \\ \hline
1 & RRR         & 3 & 2D (x,y)              & 2 & SMM-IVP, SMM Search \\ \hline
2 & xArm7       & 7 & SE3 pose              & 6 & SMM-IVP, SMM Search \\ \hline
3 & RR          & 2 & 2D (x,y) on a line    & 1 & Induced Redundancy \\ \hline
4 & xArm6       & 6 & SE3 with yaw symmetry & 5 & Induced Redundancy \\ \hline
5 & PRR         & 3 & 2D (x,y)              & 2 & Prismatic Joints \\ \hline
6 & xArm6/Rail  & 7 & SE3 pose              & 6 & Prismatic Joints \\ \hline
\end{tabular}
\end{table*}

\section{TEST CASES}\label{sec3:experiments}
We present four tests covering the SMM-IVP solver, SMM component search methods, the induced redundancy SMM-IVP, and computing SMMs for prismatic jointed systems. For each test, we provide a planar robot example and a 6/7DOF spatial manipulator example. For a planar robots, we use RR and RRR revolute jointed systems, and a PRR for a prismatic jointed model, the prismatic joint moves moves along the \(x\)-axis. All link lengths are set to 1, unless otherwise stated.

For real-world adaptability, we conducted experiments with spatial manipulators using the xarm6 and xarm7 robots from Ufactory~\cite{UFACTORYXArmUFACTORY}. We also mounted the xarm6 robot on a linear rail ($y$-axis) to simulate a prismatic jointed spatial manipulator. A summary of the test cases and robot models is presented in Table~\ref{tab:experiments}.

We used a RK5 integrator with a step size of $0.05$, and Scipy's SVD-based kernel procedure\cite{2020SciPy-NMeth}.

\section{RESULTS}\label{sec4:results}
We applied the methods presented here to compute SMM traces for an example task pose for each test case described in the previous section. Since \(k=1\) SMMs form curves in the \(n\)-dimensional configuration space, we visualize the SMM-IVP solutions using 2D plots for all \(\tfrac{1}{2}n(n-1)\) joint pair combinations in an orthographic view. Where applicable, we also provide planar robot pose plots or an image of multiple 3D spatial manipulator models superposed in the computed SMM joint configurations. The 3D view is intended to illustrate the variations in both the end-effector and base poses along the SMM solutions.
\begin{figure}[H]
    \centering
    \begin{tikzpicture}
        \node[circle,draw](X) at (-1, 0) {$\x\in\cX$};
        \node[circle,draw](Q) at (1, 0) {$\q_0\in\cC$};
        \node[circle,draw](S) at (4.5, 0) {$\cS_\x\in\cC$};
        \path[-latex] (X) edge [bend left]  node [above]  {$\phi^{-1}_{IK}$} (Q);
        \path[latex-] (X) edge [bend right] node [below]  {$\phi$} (Q);
        \path[-latex] (Q) edge  node [above]  {{\footnotesize SMM-IVP}} (S);
    \end{tikzpicture}
    \caption{SMM-IVP test cases are defined either in the task space \(\mathcal{X}\) or in the configuration space \(\mathcal{S}\). The self-motion manifold \(\mathcal{S}_\x\) is defined for \(\x = \phi(\q)\), and the initial value for the SMM-IVP solver, \(\q_0\), is obtained via inverse kinematics, i.e., \(\q_0 = \phi_{IK}^{-1}(\x)\), for examples defined in the task space.}\label{fig:test-cases-function-composition}
\end{figure}
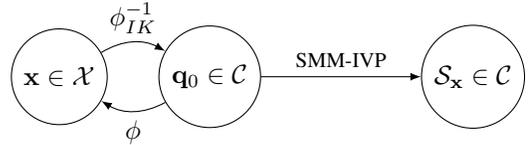
While self-motion manifolds are defined for a given task space pose \(\mathbf{x}\) (i.e., \(\mathcal{S}_\mathbf{x}\)), we present some examples starting from an initial joint configuration \(\mathbf{q}\); in such cases, we assume \(\mathbf{q}\) is obtained via some IK method for some pose \(\mathbf{x}=\phi(\mathbf{q})\). The function composition graph above shows how SMMs can be computed starting from either the task space or the configuration space.

\subsection{SMM-IVP Tests}
\subsubsection{SMM-IVP for RRR Planar Robot Model}
\begin{figure}
    \centering
    \subfigure[$\q=(-\tfrac{5\pi}{11},\tfrac{-2\pi}{3},\tfrac{\pi}{7})^\top$]
        {\includegraphics[width=0.48\linewidth]{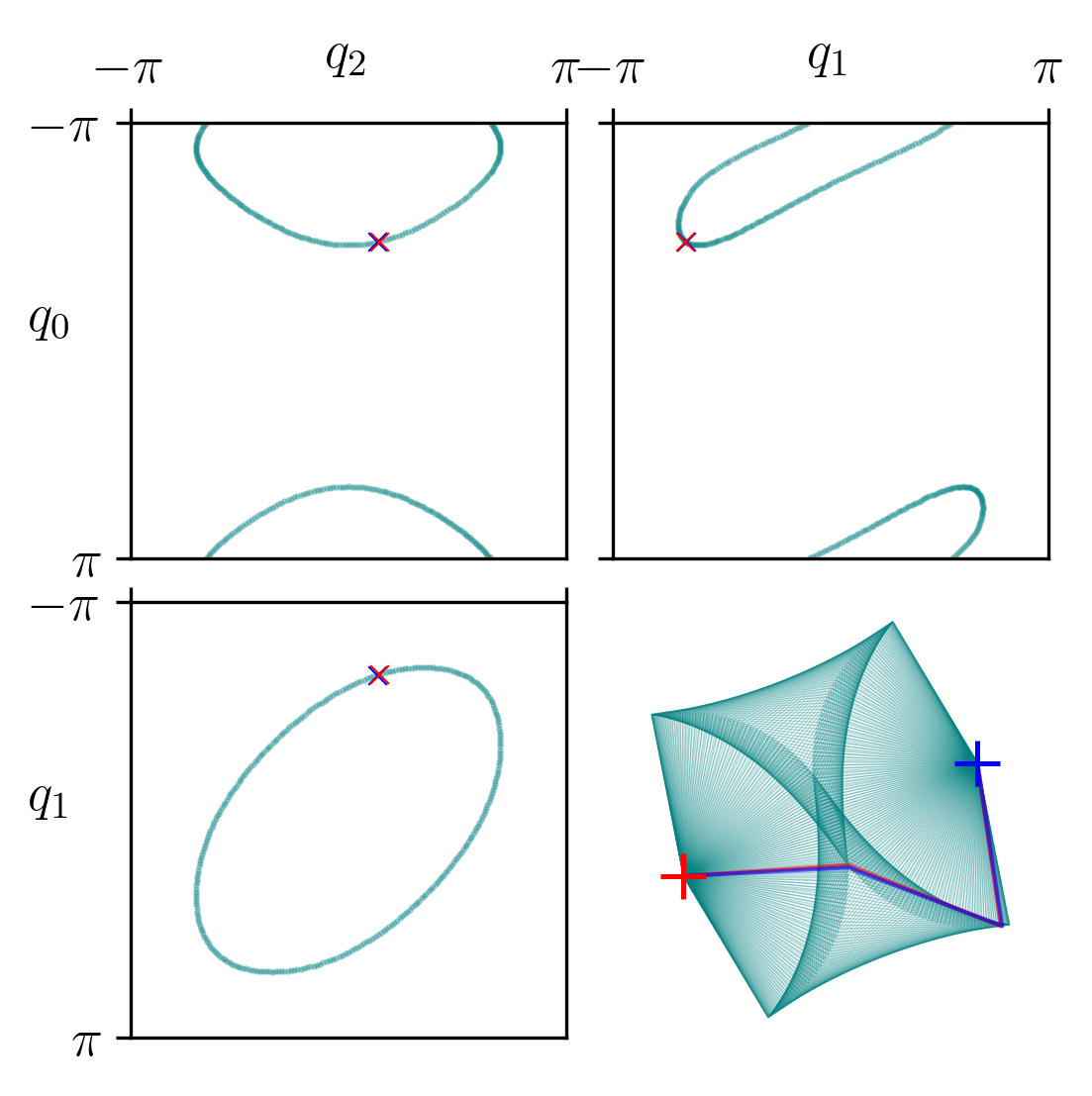}\label{fig:smm-ivp-RRR-a}}
    \subfigure[$\q=(-\tfrac{\pi}{3},\tfrac{\pi}{2},\tfrac{5\pi}{6})^\top$]
        {\includegraphics[width=0.48\linewidth]{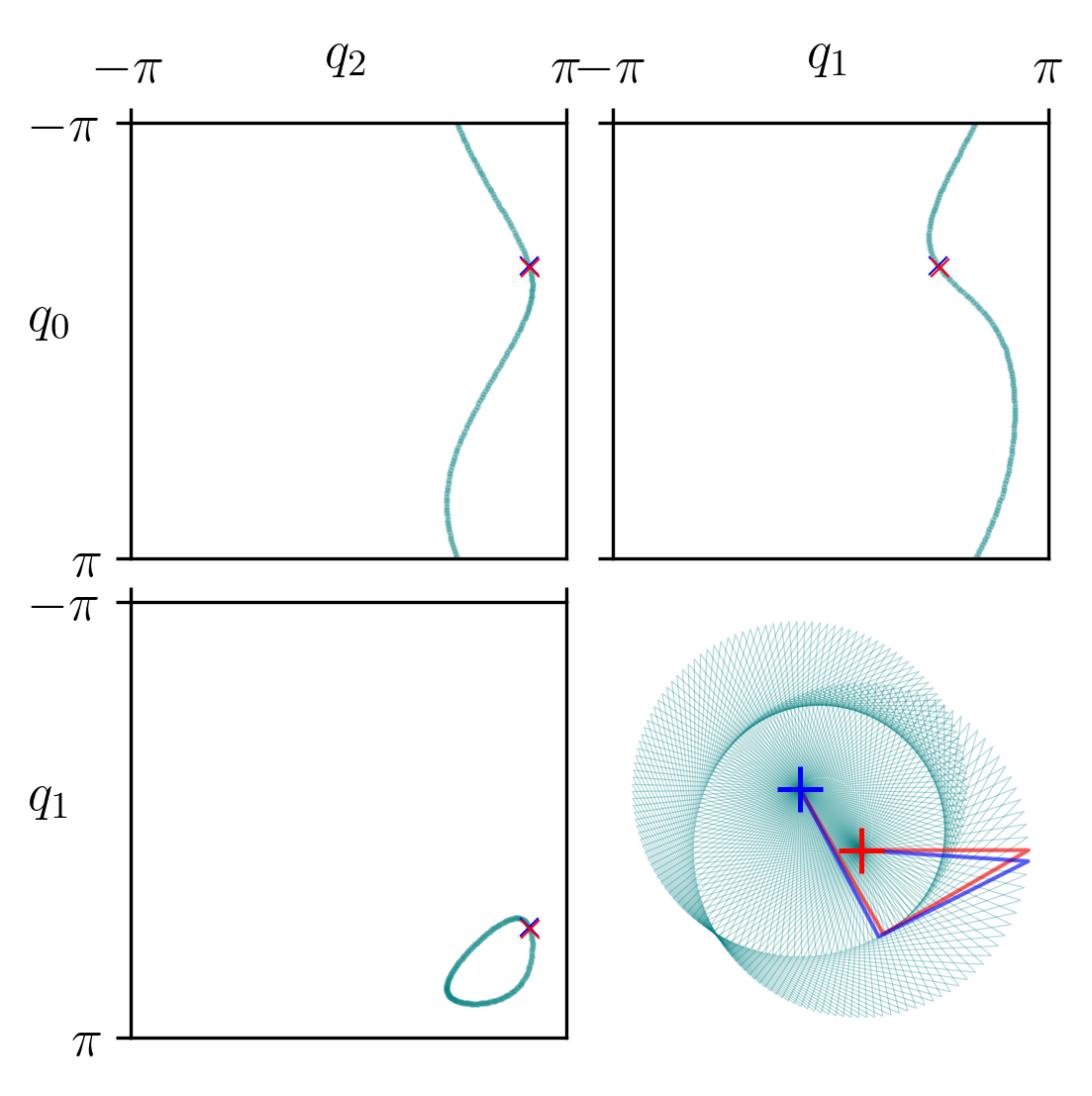}\label{fig:smm-ivp-RRR-b}}
    \caption{The figure shows two examples of SMM-IVP solutions computed for configurations \(\q\) provided above. The SMM results are shown in an orthographic view and 2D pose plots for all SMM configurations. The symbols \textcolor{red}{$\times$} and \textcolor{blue}{$\times$} indicate the initial and last SMM configurations, and the \textcolor{red}{$+$}/\textcolor{blue}{$+$}, in the 2D plots, indicate the manipulator base and end-effector.}\label{fig:smm-ivp-RRR-demos}
\end{figure}

To demonstrate the effectiveness of SMM-IVP for planar robots, we use two examples
\begin{align}
    \q=  \bigg(
        -\frac{\pi}{3},\frac{\pi}{2},\frac{5\pi}{6}
    \bigg)^\top
    \text{ and }
    \q =  \bigg(
        -\frac{5\pi}{11},\frac{-2\pi}{3},\frac{\pi}{7}
    \bigg)^\top \label{eqn:planar-example-smm-ivp}
\end{align}
defined in the configuration space. 
The examples illustrate two types of solutions the nature of which depends on the choice of the task. In Figure~\ref{fig:smm-ivp-RRR-a}, a simple closed-loop SMM is shown, which occurs when the SMM joints do not span a full $2\pi$ range of motion. In contrast, Figure~\ref{fig:smm-ivp-RRR-b} shows a case where at least one joint undergoes a $2\pi$ rotation along the SMM. In the orthographic plots, the first and last configurations of the SMM-IVP solution are marked by \textcolor{blue}{$\times$} and \textcolor{red}{$\times$}, corresponding to the blue and red markers in the 2D pose plots. Notably, the SMM-IVP solutions maintain an end-point error within $10^{-9}$ m of the initial forward kinematics end-effector position.

\subsubsection{SMM-IVP 7DOF Spatial Manipulator Example}
\begin{figure}
    \centering
    \begin{overpic}[width=0.72\linewidth]{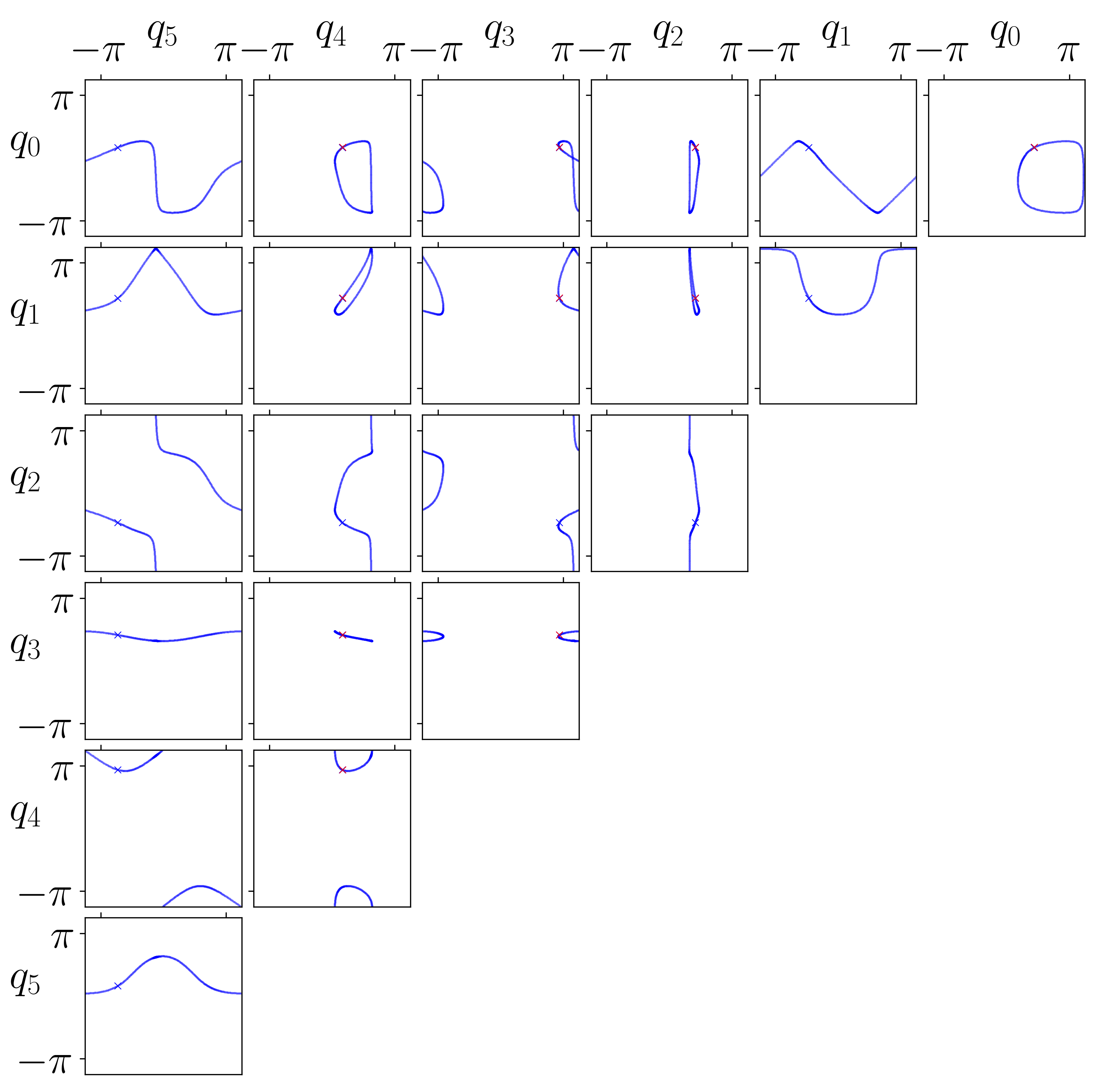}
        \put(55,0){\includegraphics[width=0.32\linewidth]{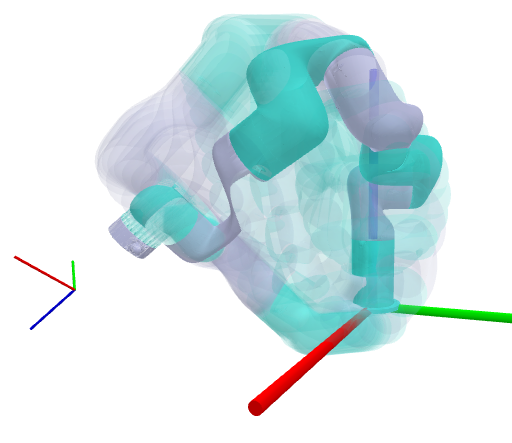}}
    \end{overpic}
    \caption{This figure shows an orthographic view of a self-motion manifold (SMM) computed using the SMM-IVP formulation of a 7DOF (xarm7) manipulator. The 3D figure shows a superposition of all SMM configurations, which demonstrates that the robot base and manipulator are constrained for all SMM configurations.}\label{fig:xarm7-smm}
\end{figure}

\begin{figure}
    \centering
    \input{results_x7smm}
    \caption{This plot shows the angle-axis error norm, $\Vert \varepsilon \Vert$, for ten SMM-IVP solutions for a 7DOF (xarm7) manipulator generated from randomly sampled joint configurations. The error for each SMM is calculated relative to the forward kinematics solution of its corresponding initial configuration. The RK5 SMM-IVP solver, using an integration step of $0.05$, achieved errors no greater than $10^{-9}$.}\label{fig:smm-ivp-error-x7}
\end{figure}
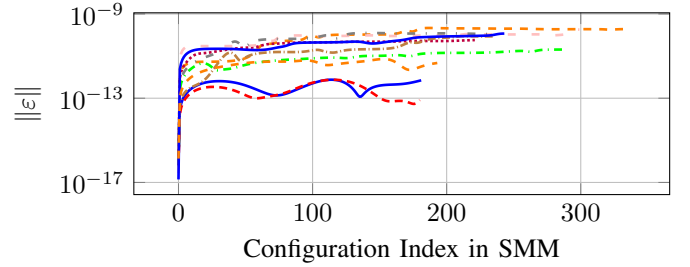

For the xarm7 manipulator, we randomly sampled within the joint limits to obtain the initial joint configuration
\[
\mathbf{q} = (0.43,\, 1.113,\, 5.098,\, 1.035,\, 2.348,\, 0.418,\, 4.468)^\top.
\]
The resulting SMM trace is presented in Figure~\ref{fig:xarm7-smm}. The 3D image qualitatively demonstrates that both the base and the end-effector remain stationary throughout the SMM-IVP solution. Quantitatively, we conducted an accuracy experiment using 10 SMM-IVP solutions for randomly sampled initial configurations. Figure~\ref{fig:smm-ivp-error-x7} shows the angle-axis errors relative to the end-effector poses of the initial configurations. In all cases, the SMM-IVP solver consistently maintained errors below \(10^{-9}\).

\subsection{Searching for Multiple Self-Motion Manifold Components}

\begin{figure}[b]
    \centering
    \subfigure[SMM Component search results for a $\mathcal{Q}^*$ set that yields one SMM component.]
    {\includegraphics[width=0.48\linewidth]{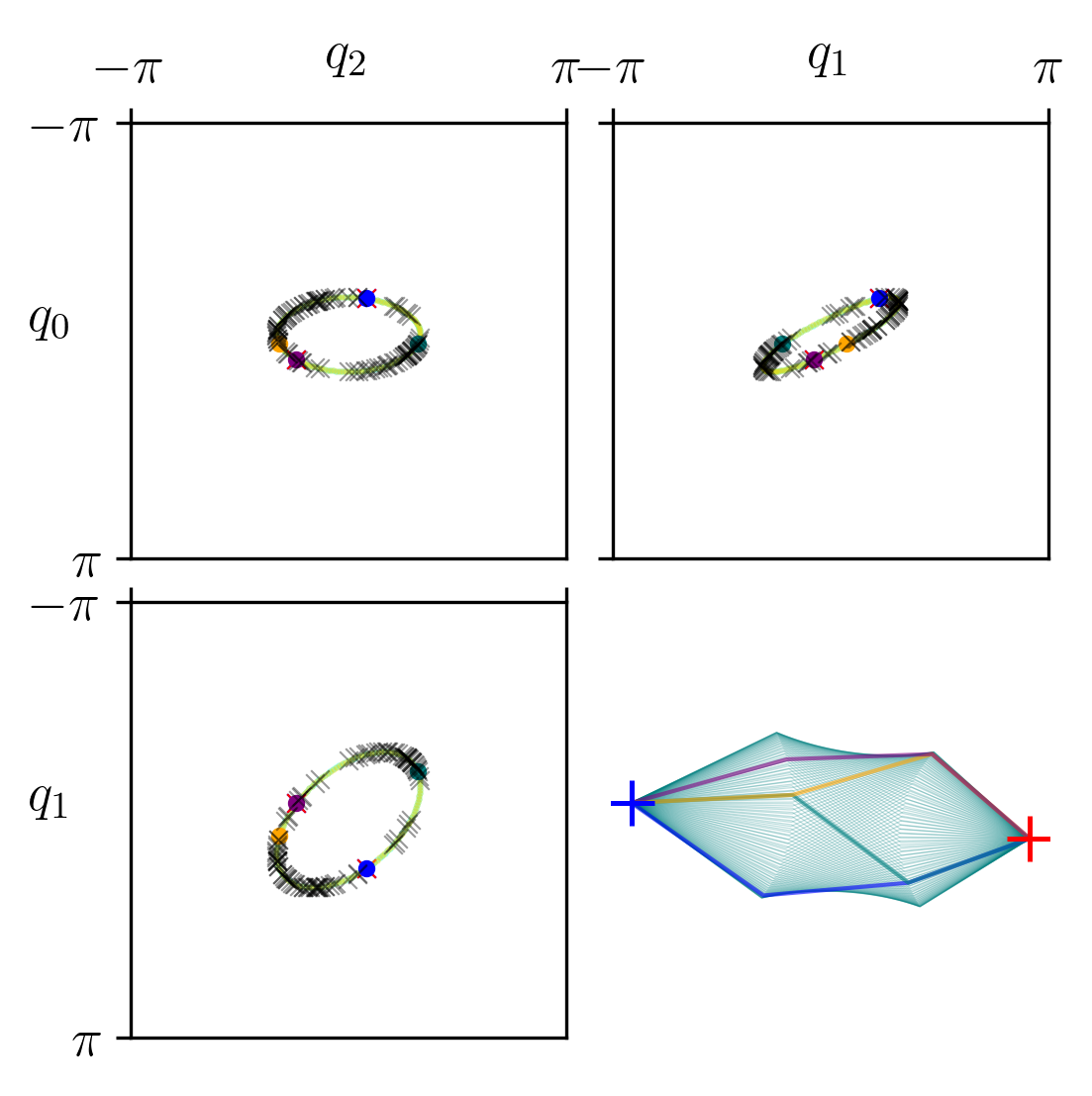}}\label{fig:toggling-one-smm}
    \subfigure[SMM Component search results for a $\mathcal{Q}^*$ set that yields two SMM component.]
    {\includegraphics[width=0.48\linewidth]{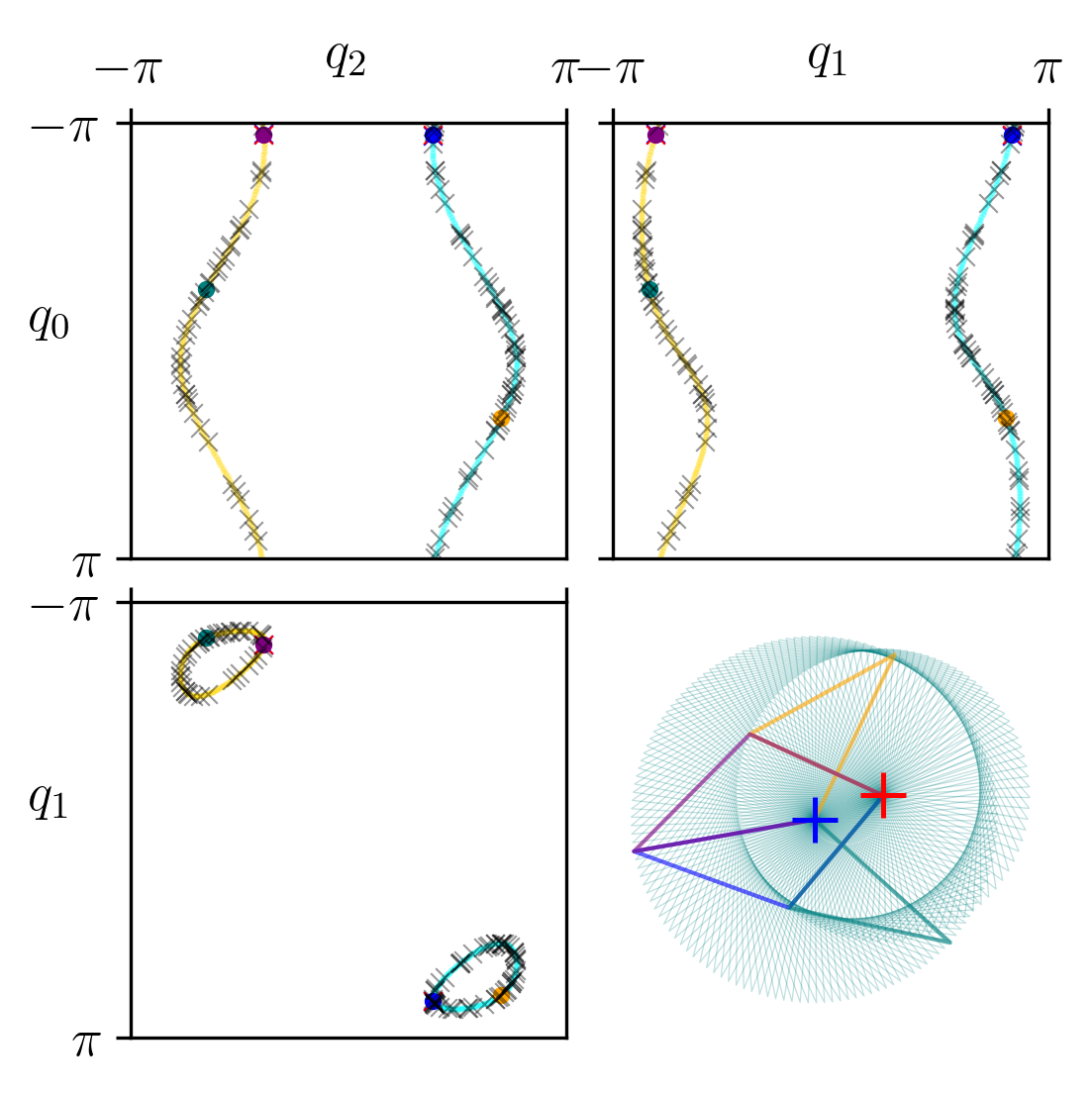}}\label{fig:toggling-two-smm}
    \caption{The results validate that the SMM components computed using the initial configuration \(\mathcal{Q}^*\) (as described in Figure~\ref{fig:toggle-illutration}) are exhaustive. For each example, 100 randomly initialized inverse kinematics (IK) solutions are plotted in the orthographic view (marked with \(\times\)). All IK solutions lie on the SMM components computed in Figure~\ref{fig:joint-toggling-smm-demo}.}\label{fig:joint-toggling-smm-search}
\end{figure}

\subsubsection{SMM component Search for RRR Planar Robot}
We demonstrate the effectiveness of the elbow-flipping approach to searching for SMM components using the examples presented in Section~\ref{sec:QstarRRR}. The examples were created from joint configurations \(\q=(-35^\circ, 40^\circ, 15^\circ)\), and \(\q=(-170^\circ,150^\circ,70^\circ)\) for and RRR robot model with lengths \(1,0.9,0.8\). 
For each example, we generated the initial configurations set $\cQ^*$ and searched for SMM components using Algorithm~\ref{alg:smm-component-search}, and the results are shown in Figure~\ref{fig:joint-toggling-smm-search}.
In Figure~\ref{fig:joint-toggling-smm-demo} we further validated these results by computing 100 random IK solutions \(\q_i=\mathsf{IK}(\phi_{{}_{3R}}(\q))\). All 100 IK solutions (marked with $\times$) lie on the SMM components obtained using $\cQ^*$ and Algorithm~\ref{alg:smm-component-search}. We take this as confirmation that the proposed $\cQ^*$ efficiently generates a set that guarantees exploration of all available SMM components.
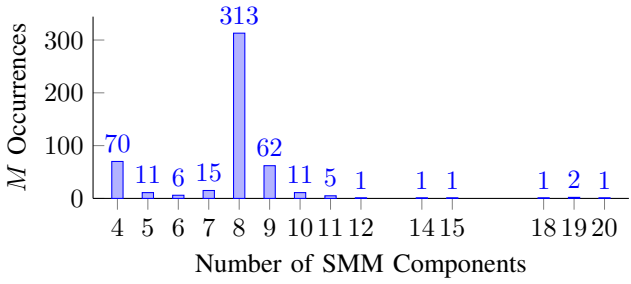
\begin{figure}
    \centering
    \begin{tikzpicture}
        \begin{axis}[
            ybar, 
            xtick=data, 
            height=4cm,
            width=0.48\textwidth,
            axis x line*=bottom,
            axis y line*=left,
            ylabel={$M$ Occurrences},
            xlabel={Number of SMM Components},
            ymin=0, 
            bar width=0.15cm, 
            nodes near coords, 
            enlarge x limits=0.05 
        ]
        \addplot table[x=Multiplicity,y=Occurrences,col sep=comma] {results/x7smm/multiplicity.csv}; 
        \end{axis}
    \end{tikzpicture}
    \caption{The plot shows the distribution of the number of SMM components for 500 randomly sampled xarm7 configurations. 90\% of the xArm7 manipulator's workspace is characterized by SMM components {$4$}, {$8$}, and {$9$}; a naive gradient-based IK solver arbitrarily lands on one of the components by chance. Excessively high SMM component counts are due to sampling near workspace surface singularities where the SMM-IVP struggles.}\label{fig:x7-smm-components}
\end{figure}

\subsubsection{SMM component Search for a 7DOF Spatial Manipulator}
For the xarm7 spatial manipulator, we used the initial configurations set $\cQ^*$ with $N=150$ randomly initialized IK solutions for a selected task $\x$. From prior tests, the number of SMM components converged within $N=30$. However, we settled for $150$ to increase the likelihood of convergence. We applied this SMM components search approach to  500 end-effector poses computed from randomly sampled task poses. Figure~\ref{fig:x7-smm-components} shows the distribution of SMM components across the 500 poses. About \(60\%\) of the robot's workspace is characterized by 8 SMM components, and components 4, 8, and 9 constitute \(90\%\) of the robot's workspace.
\begin{figure}
    \centering
    \begin{tikzpicture}
        \begin{axis}[
            ybar, 
            xtick=data, 
            height=4cm,
            width=0.48\textwidth,
            axis x line*=bottom,
            axis y line*=left,
            ylabel={Occurrences},
            xlabel={Number of Usable SMM Components},
            ymin=0, 
            bar width=0.15cm, 
            nodes near coords, 
            enlarge x limits=0.05 
        ]
        \addplot table[x=Multiplicity,y=Occurrences,col sep=comma] {results/x7smm/UsefulMultiplicities.csv}; 
        \end{axis}
    \end{tikzpicture}
    \caption{This plot presents the distribution of usable SMM components from the data in Figure~\ref{fig:x7-smm-components}. Usable SMM components are defined as SMM-IVP solutions with at least one configuration within joint limits, i.e., usable. The results show that $92\%$ of the xarm7 workspace is characterized by $2$, $4$, or $5$ usable SMM component counts. Excessively high SMM component counts are due to SMM-IVP ineffectiveness near singularities.}\label{fig:x7-usable-smm-components}
\end{figure}
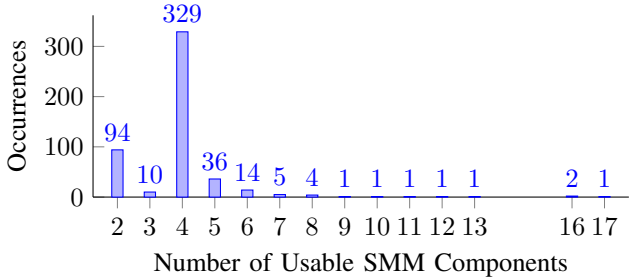
A significant number of components lie entirely outside the xarm7's joint limits. Figure~\ref{fig:x7-usable-smm-components} presents the distribution of \textit{usable SMM components} --- those that contain at least one configuration within the joint limits. Slightly over \(60\%\) of the workspace has 4 usable components, and {\(92\%\) is comprised of 2, 4, or 5 components}.

\begin{figure}
    \centering
    \subfigure[$\;$]{\includegraphics[width=0.24\linewidth,valign=t]{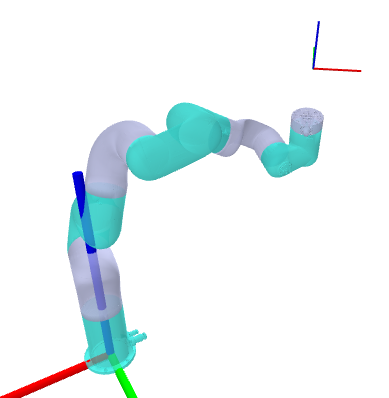}}\label{fig:a}
    \subfigure[$\;$]{\includegraphics[width=0.24\linewidth,valign=t]{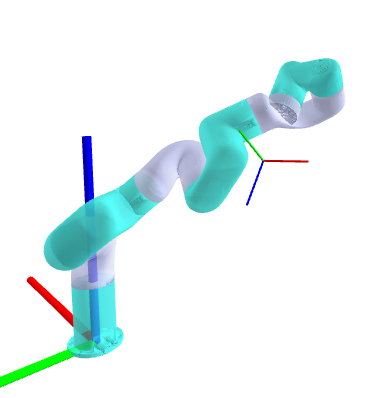}}\label{fig:b}
    \subfigure[$\;$]{\includegraphics[width=0.24\linewidth,valign=t]{images/b19a.png}}\label{fig:c}
    \subfigure[$\;$]{\includegraphics[width=0.24\linewidth,valign=t]{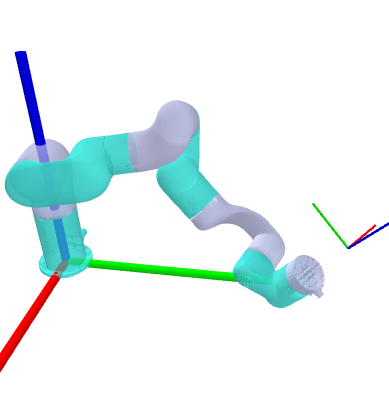}}\label{fig:d}
    \caption{The robot poses above correspond with the SMM component counts 18, 19, 19, and 20 in Figure~\ref{fig:x7-smm-components}. These are poses that are near the limits of the robot's reach, where the SMM-IVP method is characterized by poor exits and distance checking. More details in Table.~\ref{tab:spurious-components-poses}.}\label{fig:spurious-smm-counts}
\end{figure}

The data in Figures~\ref{fig:x7-smm-components} and~\ref{fig:x7-usable-smm-components} shows numerous end-effector poses that appear to generate more than 16 SMM components, which would seem to violate Burdick's hypothesis. However, as extensively discussed by Peidr\'o~\etal~\cite{peidroMethodBasedVanishing2018}, and also consistent with low manipulability on and near barrier singularities~\cite{bohigasCompleteMethodWorkspace2012}, these cases (shown in Figure~\ref{fig:spurious-smm-counts} and Table~\ref{tab:spurious-components-poses}) occur near barrier singularities where the SMM-IVP methods yield unreliable results because the Jacobian is ill-conditioned and the SMM is collapsing into a single point\cite{peidroMethodBasedVanishing2018}. Therefore, these SMM component counts should not be considered as actual violations of the hypothesis.

\begin{table}
    \centering
    \caption{The poses below (also presented in Figure~\ref{fig:spurious-smm-counts}) resulted in excessively high SMM component counts in Figure~\ref{fig:x7-smm-components}.}\label{tab:spurious-components-poses}
    \begin{tabular}{|l|c|c|} \hline
        Sample  & \textbf{Position}  & \textbf{Orientation (RPY/degrees)} \\ \hline
        (a) & $(-0.497, 0.111, 0.668)$  & $(17.3, 1.2, 139)$    \\ \hline
        (b) & $(-0.381, -0.195, 0.464)$ & $(138, -3.92, -166)$  \\ \hline
        (c) & $(-0.00295, -0.23, 1)$    & $(13.6, 18, -158)$    \\ \hline
        (d) & $(0.131, 0.621, 0.211)$   & $(34.7, -0.0794, 143)$\\ \hline
\end{tabular}
\end{table}

\begin{figure}
    \centering
        \begin{overpic}[width=0.72\linewidth]{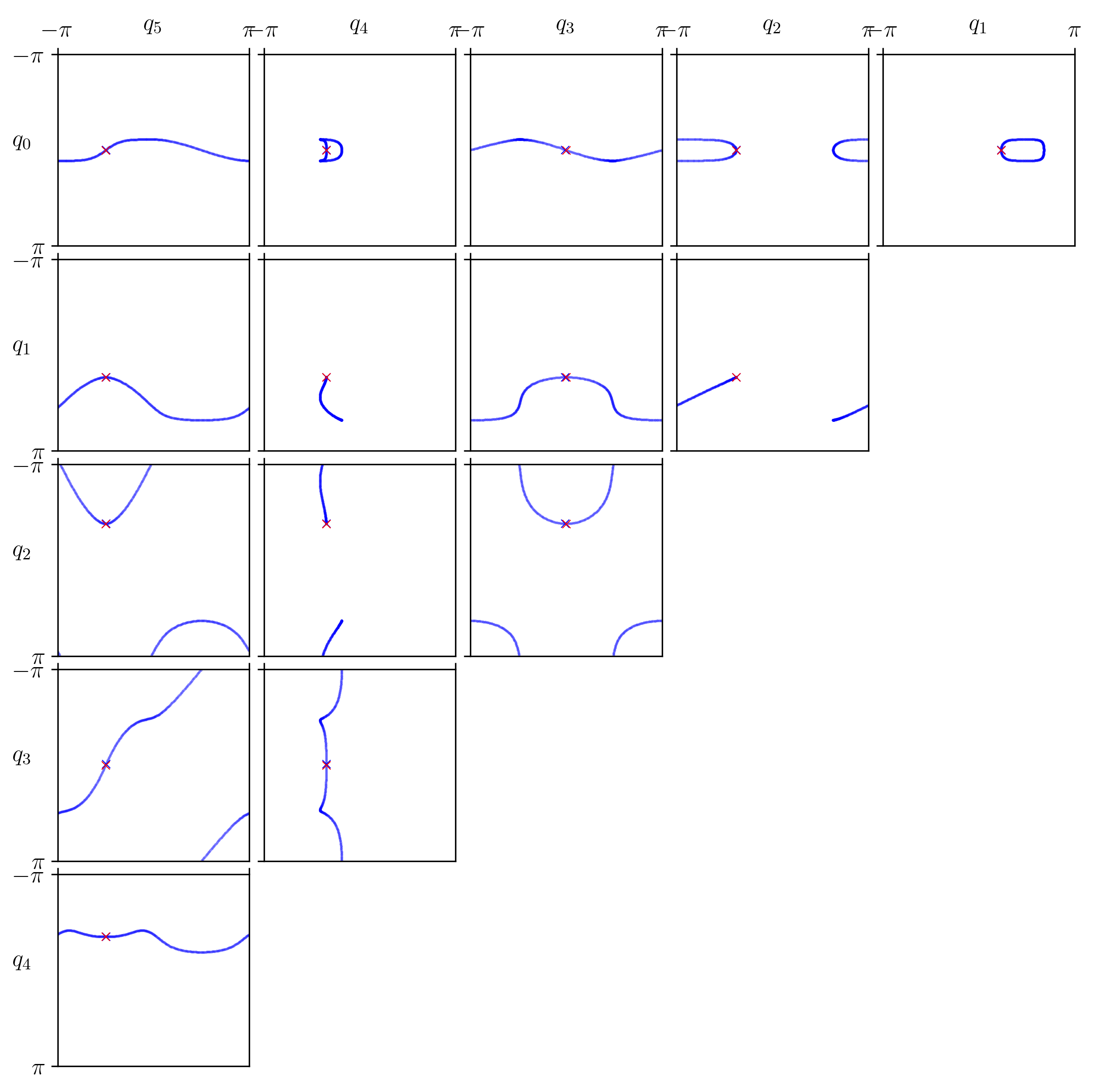}
            \put(62,2){\includegraphics[width=0.5\linewidth]{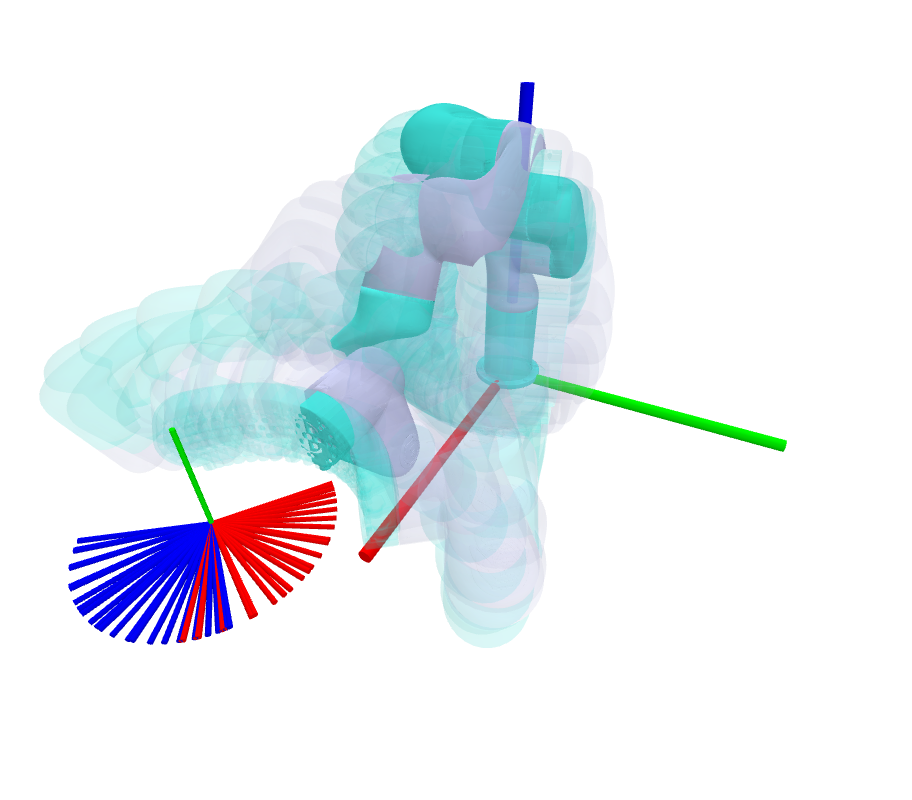}}
        \end{overpic}
    \caption{This figure shows an orthographic view of the self-motion manifold of a 6DOF manipulator assigned to a 5DOF task. The SMM is computed using the induced redundancy framework presented in Section~\ref{sec2:problem:induced}. The 3D view is a superposition of the SMM poses --- it demonstrates the rotation about the end-effector's $y$-axis as induced by the redundancy vector $\mathbf{u}=(0,0,0,0,1,0)^\top$.}\label{fig:xarm6-smm}
\end{figure}

\subsection{Induced Redundancy SMM-IVP}\label{sec2:problem:induced}

\subsubsection{Induced Redundancy for an RR Planar Robot}
\begin{figure}
    \centering
    \subfigure[$\mathbf{u}=(1, 0.5)$]
    {
        \includegraphics[width=0.24\linewidth,valign=t]{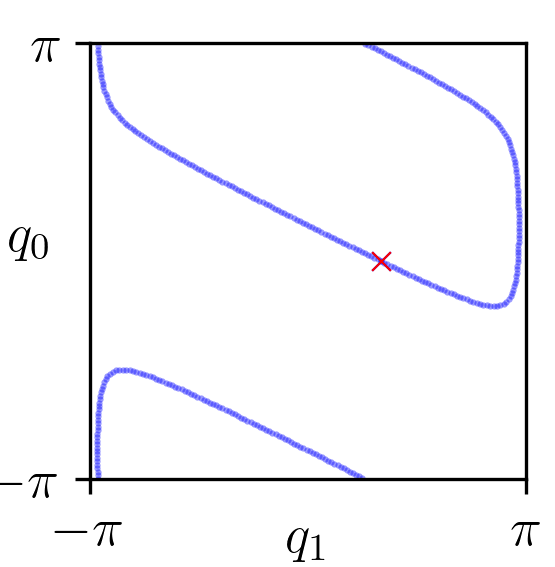}
        \includegraphics[width=0.24\linewidth,valign=t]{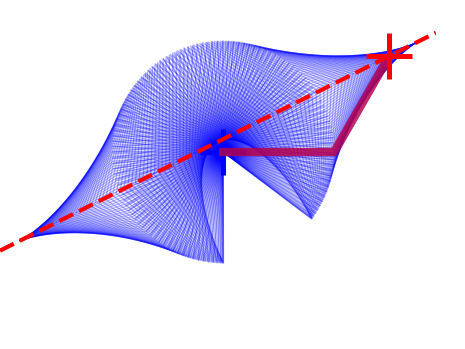}
    }\label{fig:toggling-two-smm}
    \subfigure[$\mathbf{u}=(-0.5, 1)$]
    {
        \includegraphics[width=0.24\linewidth,valign=t]{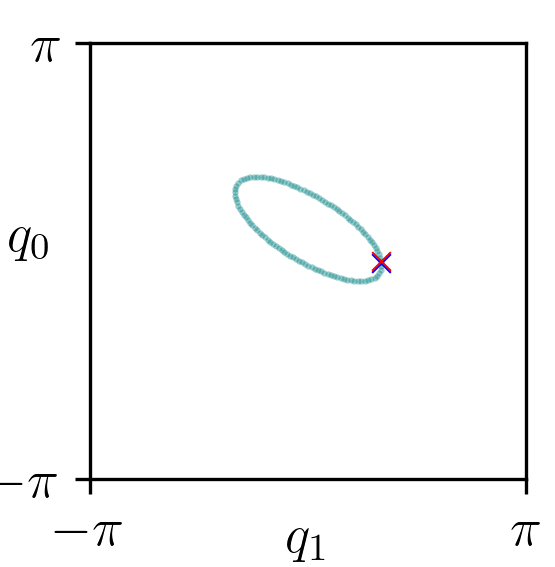}
        \includegraphics[width=0.18\linewidth,valign=t]{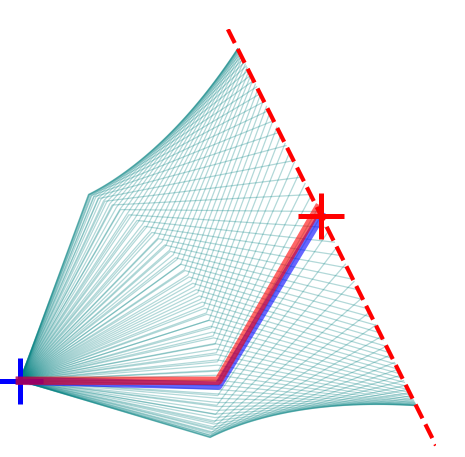}
    }\label{fig:toggling-two-smm}
    \caption{The induced redundancy SMM-IVP solutions above were computed for the RR planar model starting from \(\mathbf{q}=(0, \tfrac{\pi}{3})^\top\), using the redundancy directions specified in the respective captions. The results demonstrate that the solver successfully tracks the natural 1DOF representation of the task, as defined by the redundancy direction \(\mathbf{u}\).}\label{fig:smm-ivp-u2R-demos}
\end{figure}

The planar example for induced redundancy SMM-IVP was applied to the RR robot, a non-redundant model. We arbitrarily initialized the model at \(\mathbf{q}=(0, \tfrac{\pi}{3})^\top\) and selected two perpendicular redundancy directions, \(\mathbf{u}_0=(1,0.5)^\top\) and \(\mathbf{u}_1=(-0.5,1)^\top\). In the induced redundancy SMM-IVP approach, the redundancy directions \(\mathbf{u}_i\) are first normalized to \(\hat{\mathbf{u}}_i\) before computing the projection matrix. Figure~\ref{fig:smm-ivp-u2R-demos} presents the results for both directions --- the \(\mathbf{u}_0\) solution in blue and the one for \(\mathbf{u}_1\) in teal. The first plot shows the SMM traces in the joint space, with the initial and final configurations marked by \textcolor{red}{\(\times\)} and \textcolor{blue}{\(\times\)}, respectively. The second and third plots display the corresponding 2D pose plots for the \(\mathbf{u}_0\) and \(\mathbf{u}_1\) solutions, with the red dashed line indicating the subspace defined by each redundancy direction. In both cases, the SMM-IVP solver successfully tracked the specified subspace and terminated correctly.

\subsubsection{Induced Redundancy for a 6DOF Spatial Manipulator}

\begin{figure}
    \centering
    \subfigure[End-effector position error $\Vert \delta \Vert$]
    {\input{results_x6smm_a}\label{fig:smm-ivp-error-x6-position}}
    \\
    \subfigure[End-effector tool frame y-axis angular deflection error $\Vert \gamma \Vert$.]
    {\input{results_x6smm_b}\label{fig:smm-ivp-error-x6-y-axis}}
    \caption{The plots above show the position error \(\Vert \delta \Vert\) and angular deflection error \(\Vert \gamma \Vert\) for an induced redundancy SMM-IVP with \(\mathbf{u}=(0,0,0,0,1,0)^\top\), computed from ten randomly sampled configurations. Both error measures, calculated relative to the initial joint configuration’s end-effector pose, do not exceed \(10^{-7}\).}\label{fig:smm-ivp-error-x6}
\end{figure}
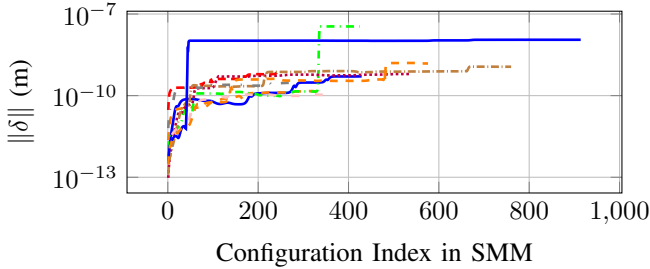
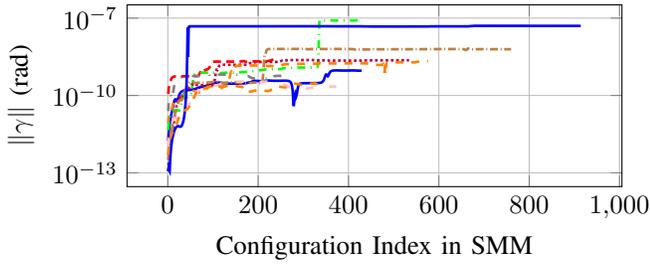

In this example, we chose a redundancy direction \(\mathbf{u}=(0,0,0,0,1,0)^\top\) defined in the end-effector frame (i.e., using the tool-frame Jacobian in Equation~\ref{eqn:ir-regularized-ode}) to model a grasping task where the \(y\)-axis is perpendicular to the grasping plane.
The induced redundancy SMM-IVP was applied to xarm6 spatial manipulator, starting from joint configuration configuration 
\[
\mathbf{q} = (0,\; 0.379,\; -1.17,\; -3.142,\; 0.779,\; 1.571)^\top.
\]
The resulting SMM-IVP solution is presented Figure~\ref{fig:xarm6-smm}. The 3D image illustrates that the end-effector rotates about the \(y\)-axis without any change in position. Figures~\ref{fig:smm-ivp-error-x6-position} and~\ref{fig:smm-ivp-error-x6-y-axis} present the position and \(y\)-axis angular deflection errors, respectively, for 10 induced redundancy SMM-IVP solutions computed for randomly sampled joint configurations with \(\mathbf{u}=(0,0,0,0,1,0)^\top\). In all cases, the SMM-IVP solutions maintain errors consistently below \(10^{-7}\) for both position and \(y\)-axis angular deflection.

\subsection{SMM-IVP for Prismatic Jointed Robot Systems}

\subsubsection{SMM-IVP for PRR Planar Robot Model}
\begin{figure}
    \centering
    \subfigure[$\q=(-\frac{1}{2}, 0, \frac{\pi}{4})^\top$]
            {\includegraphics[width=0.48\linewidth]{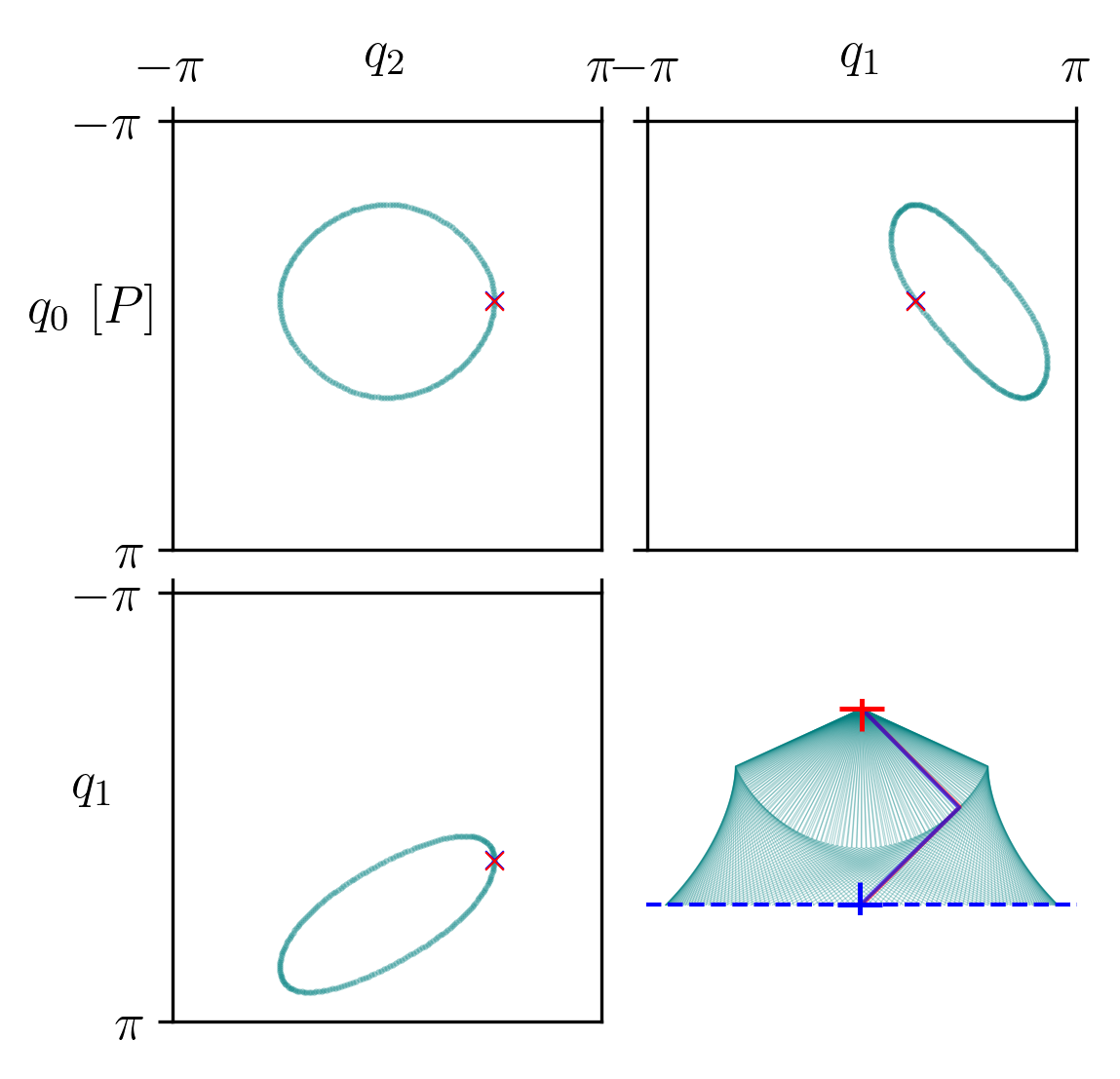}\label{fig:smm-ivp-PRR-a}}
    \subfigure[$\q=(-\frac{1}{2}, \frac{\pi}{4}, -\frac{\pi}{2})^\top$]
        {\includegraphics[width=0.48\linewidth]{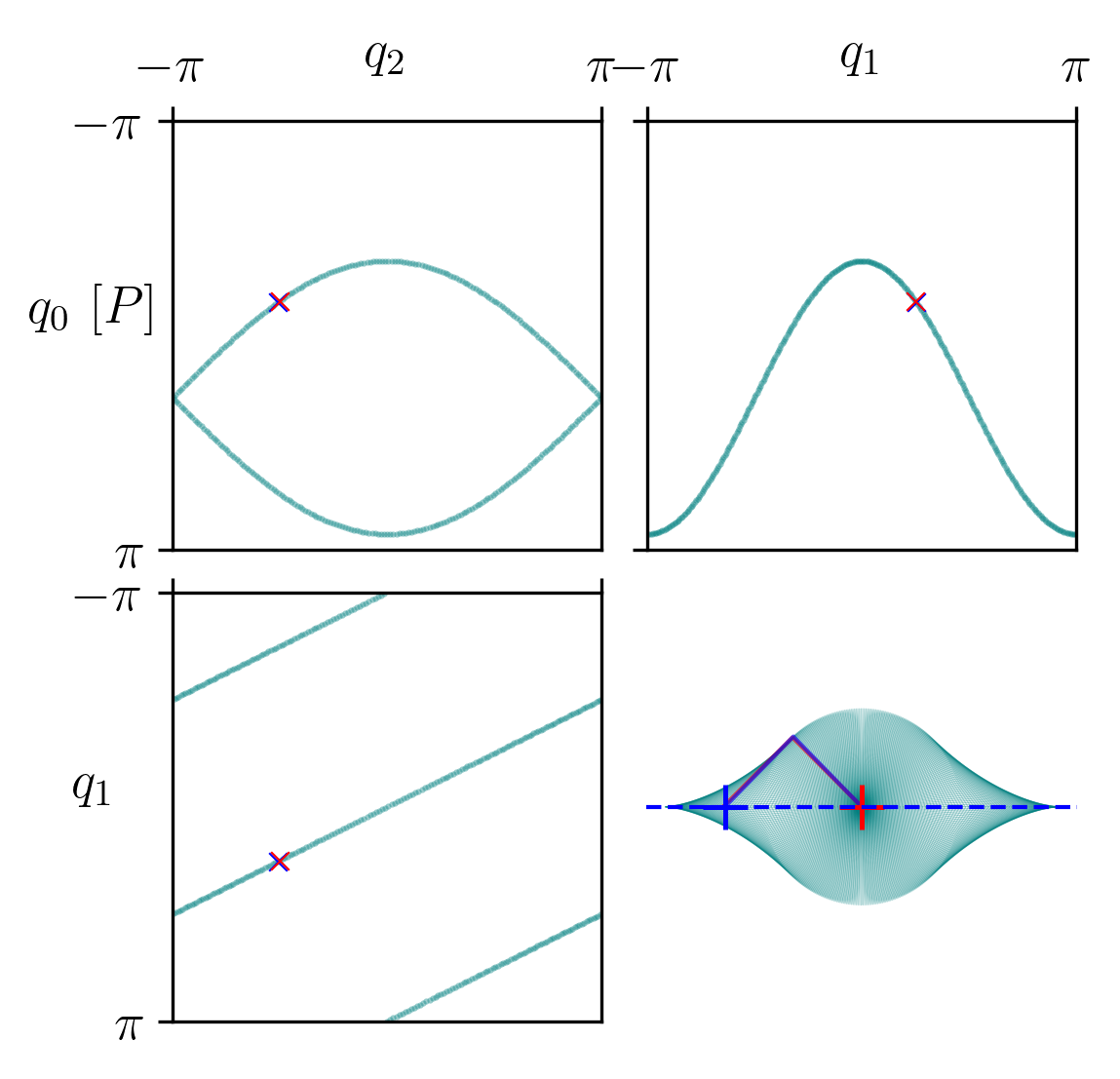}\label{fig:smm-ivp-PRR-b}}
    \caption{The plots above show two SMM examples for a prismatic jointed PRR planar robot. In both cases, the SMM-IVP solver successfully tracks and exits a self-motion manifold. The first joint, labeled $q_0~[P]$, translates along the $x$-axis (blue dashed line). The symbols \textcolor{red}{$\times$} and \textcolor{blue}{$\times$} indicate the initial and last joint configurations, and \textcolor{red}{$+$} marks the target end-effector position.}\label{fig:smm-ivp-PRR}
\end{figure}

Figure~\ref{fig:smm-ivp-PRR} shows two examples of SMM traces computed using the SMM-IVP method for the PRR planar robot model. The examples use the initial configurations
\begin{align*}
    \q=  \bigg(
        -\frac{1}{2}, 0, \frac{\pi}{4}
    \bigg)^\top
    \text{ and }
    \q =  \bigg(
        -\frac{1}{2}, \frac{\pi}{4}, -\frac{\pi}{2}
    \bigg)^\top
\end{align*}
which were chosen to demonstrate two distinctly different cases of SMMs in a PRR model. In both cases, the SMM-IVP solutions remain constrained to the \(x\)-axis (indicated by the blue dashed line). The method starts from the initial configuration, successfully traverses the SMM, and returns correctly to terminate.

\begin{figure}[b]
    \centering
        \begin{overpic}[width=0.72\linewidth]{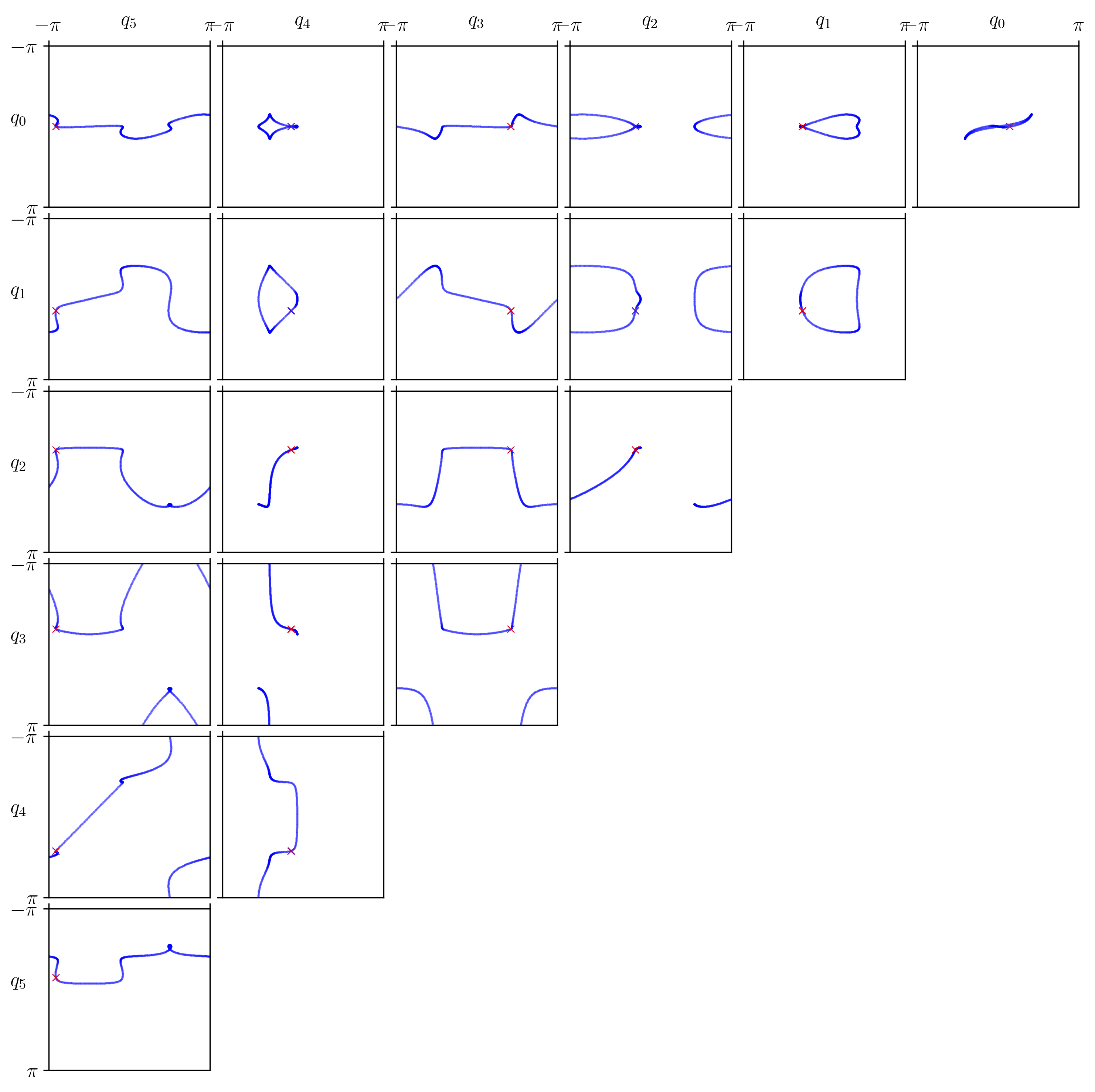}
            \put(40,-20){\includegraphics[width=0.72\linewidth]{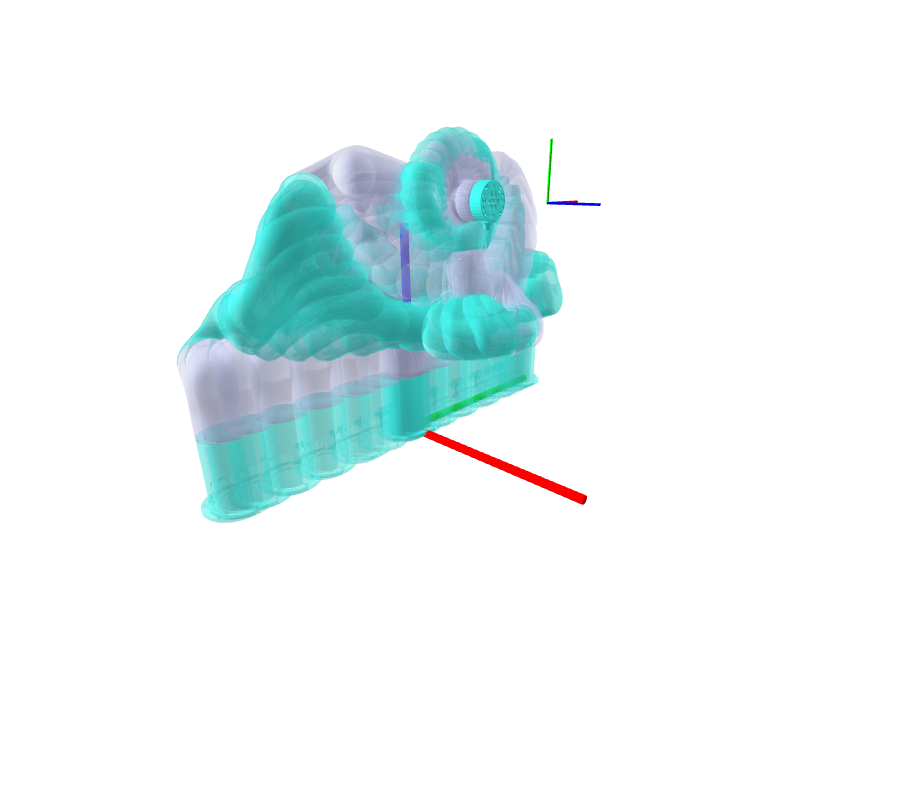}}
        \end{overpic}
    \caption{The SMM results above are for a prismatic jointed 7DOF spatial manipulator (xarm6 on \(y\)-axis linear rail). The SMM-IVP correctly tracks the linear motion and successfully exits upon returning to the initial configuration. This result demonstrates that the SMM-IVP can be applied to prismatic jointed systems without any modifications.}\label{fig:xarm6-smm-t}
\end{figure}
\subsubsection{SMM-IVP for a Prismatic Jointed 7DOF Spatial Manipulator}
The SMM-IVP implementation also worked for the xarm6 spatial manipulator mounted on a $y$-axis linear rail. The results in Figure~\ref{fig:xarm6-smm} show a superposition of some SMM configurations that demonstrates that the robot slides along the $y$-axis but still maintains the same end-effector pose.

\section{LIMITATIONS}
The SMM-IVP formulation suffers from two significant limitations, both caused by singular configurations. The first is using singular configurations as initial values. Because the Jacobian kernel is multi-dimensional, i.e., \(k>1\), which violates the \(k=1\) assumption required for the directionally regularized ODE, Equation~\ref{eqn:regularized-ode}. In practice, a multi-dimensional null space makes determining a unique heading direction impossible. However, this limitation is overcome by solving the IK problem for different configuration.

The second limitation arises near barrier singularities~\cite{bohigasCompleteMethodWorkspace2012}, where self-motion manifolds progressively shrink and eventually collapse to a point~\cite{peidroMethodBasedVanishing2018}. Under these conditions, we attribute the poor SMM-IVP solver performance to an overly large integration step \(h\), which causes the solver to overstep the manifold and prevents it from tracing back to the initial configuration for proper termination.

\section{CONCLUSION}\label{sec5:conclusion}

In this work, we introduced the self-motion manifold initial value problem (SMM-IVP) ODE formulation for computing self-motion manifolds (SMMs) and its induced redundancy variant for non-redundant manipulators assigned to lower-dimensional tasks. The SMM-IVP formulation simplifies SMM computation by employing higher-order explicit ODE solvers that yield highly accurate joint configurations without requiring additional inverse kinematics refinement --- a benefit we attribute to the ODE solver quality. We further integrate the SMM-IVP formulation into a rejection sampling-based algorithm for efficient SMM component search. We also demonstrated that our methods work for prismatic jointed systems, even though SMM theory has traditionally been developed for revolute manipulators.

Future work will focus on enhancing the computation of \(k=1\) SMMs near workspace surface singularities and on developing approaches for higher-order redundancy (i.e., \(k>1\)) self-motion manifolds. We also intend to explore the integration of SMMs into motion planning and trajectory optimization.

\section*{ACKNOWLEDGMENT}
This work was supported in part by NSF Robust Intelligence 1956163, NSF/USDA-NIFA AIIRA AI Research Institute 2021-67021-35329, and USDA-NIFA LEAP 2024-51181-43291.


\balance
\bibliographystyle{IEEEtran}
\bibliography{structured-design}

\end{document}

%% file: results_x7smm.tex
\begin{tikzpicture}
    \begin{semilogyaxis}[
        cycle list name=colorblind-friendly, 
        xlabel={Configuration Index in SMM},
        ylabel={$\Vert \varepsilon \Vert$},
        grid=major,
        width=0.48\textwidth,
        height=4cm,
        ]

        \addplot table [
            col sep=comma,
            x=Index,
            y=Error,
        ] {results/x7smm/angle_axis_error_00.csv};

        \addplot table [
            col sep=comma,
            x=Index,
            y=Error,
        ] {results/x7smm/angle_axis_error_01.csv};
        
        \addplot table [
            col sep=comma,
            x=Index,
            y=Error,
        ] {results/x7smm/angle_axis_error_02.csv};
        
        \addplot table [
            col sep=comma,
            x=Index,
            y=Error,
        ] {results/x7smm/angle_axis_error_03.csv};
        
        \addplot table [
            col sep=comma,
            x=Index,
            y=Error,
        ] {results/x7smm/angle_axis_error_04.csv};
        
        \addplot table [
            col sep=comma,
            x=Index,
            y=Error,
        ] {results/x7smm/angle_axis_error_05.csv};
        
        \addplot table [
            col sep=comma,
            x=Index,
            y=Error,
        ] {results/x7smm/angle_axis_error_06.csv};
        
        \addplot table [
            col sep=comma,
            x=Index,
            y=Error,
        ] {results/x7smm/angle_axis_error_07.csv};
        
        \addplot table [
            col sep=comma,
            x=Index,
            y=Error,
        ] {results/x7smm/angle_axis_error_08.csv};
        
        \addplot table [
            col sep=comma,
            x=Index,
            y=Error,
        ] {results/x7smm/angle_axis_error_09.csv};
        
    \end{semilogyaxis}
\end{tikzpicture}

%% file: results_x6smm_a.tex
\begin{tikzpicture}
    \begin{semilogyaxis}[
        cycle list name=colorblind-friendly, 
        xlabel={Configuration Index in SMM},
        ylabel={$\Vert \delta \Vert$ (m)},
        grid=major,
        width=0.45\textwidth,
        height=4cm,
        legend pos = outer north east
        ]

        \addplot table [
            col sep=comma,        
            x=Index,              
            y=DistanceError       
        ] {results/x6smm/distance_ydeflection_error_00.csv};   
        \addplot table [
            col sep=comma,        
            x=Index,              
            y=DistanceError       
        ] {results/x6smm/distance_ydeflection_error_01.csv};   
        \addplot table [
            col sep=comma,        
            x=Index,              
            y=DistanceError       
        ] {results/x6smm/distance_ydeflection_error_02.csv};   
        \addplot table [
            col sep=comma,        
            x=Index,              
            y=DistanceError       
        ] {results/x6smm/distance_ydeflection_error_03.csv};   
        \addplot table [
            col sep=comma,        
            x=Index,              
            y=DistanceError       
        ] {results/x6smm/distance_ydeflection_error_04.csv};
        
        \addplot table [
            col sep=comma,        
            x=Index,              
            y=DistanceError       
        ] {results/x6smm/distance_ydeflection_error_05.csv};
        \addplot table [
            col sep=comma,        
            x=Index,              
            y=DistanceError       
        ] {results/x6smm/distance_ydeflection_error_06.csv};
        \addplot table [
            col sep=comma,        
            x=Index,              
            y=DistanceError       
        ] {results/x6smm/distance_ydeflection_error_07.csv};
        \addplot table [
            col sep=comma,        
            x=Index,              
            y=DistanceError       
        ] {results/x6smm/distance_ydeflection_error_08.csv};
        \addplot table [
            col sep=comma,        
            x=Index,              
            y=DistanceError       
        ] {results/x6smm/distance_ydeflection_error_09.csv};
        
    \end{semilogyaxis}
\end{tikzpicture}

%% file: results_x6smm_b.tex
\begin{tikzpicture}
    \begin{semilogyaxis}[
        cycle list name=colorblind-friendly, 
        xlabel={Configuration Index in SMM},
        ylabel={$\Vert \gamma \Vert$ (rad)},
        grid=major,
        width=0.45\textwidth,
        height=4cm,
        legend pos = outer north east
        ]

        \addplot table [
            col sep=comma,        
            x=Index,              
            y=YDirectionError       
        ] {results/x6smm/distance_ydeflection_error_00.csv};   
        \addplot table [
            col sep=comma,        
            x=Index,              
            y=YDirectionError       
        ] {results/x6smm/distance_ydeflection_error_01.csv};   
        \addplot table [
            col sep=comma,        
            x=Index,              
            y=YDirectionError       
        ] {results/x6smm/distance_ydeflection_error_02.csv};   
        \addplot table [
            col sep=comma,        
            x=Index,              
            y=YDirectionError       
        ] {results/x6smm/distance_ydeflection_error_03.csv};   
        
        \addplot table [
            col sep=comma,        
            x=Index,              
            y=YDirectionError       
        ] {results/x6smm/distance_ydeflection_error_04.csv};

        \addplot table [
            col sep=comma,        
            x=Index,              
            y=YDirectionError       
        ] {results/x6smm/distance_ydeflection_error_05.csv};
        
        \addplot table [
            col sep=comma,        
            x=Index,              
            y=YDirectionError       
        ] {results/x6smm/distance_ydeflection_error_06.csv};
        
        \addplot table [
            col sep=comma,        
            x=Index,              
            y=YDirectionError       
        ] {results/x6smm/distance_ydeflection_error_07.csv};
        
        \addplot table [
            col sep=comma,        
            x=Index,              
            y=YDirectionError       
        ] {results/x6smm/distance_ydeflection_error_08.csv};
        
        \addplot table [
            col sep=comma,        
            x=Index,              
            y=YDirectionError       
        ] {results/x6smm/distance_ydeflection_error_09.csv};
    \end{semilogyaxis}
\end{tikzpicture}

%% file: main.bbl
\begin{thebibliography}{10}
\providecommand{\url}[1]{#1}
\csname url@samestyle\endcsname
\providecommand{\newblock}{\relax}
\providecommand{\bibinfo}[2]{#2}
\providecommand{\BIBentrySTDinterwordspacing}{\spaceskip=0pt\relax}
\providecommand{\BIBentryALTinterwordstretchfactor}{4}
\providecommand{\BIBentryALTinterwordspacing}{\spaceskip=\fontdimen2\font plus
\BIBentryALTinterwordstretchfactor\fontdimen3\font minus
  \fontdimen4\font\relax}
\providecommand{\BIBforeignlanguage}[2]{{%
\expandafter\ifx\csname l@#1\endcsname\relax
\typeout{** WARNING: IEEEtran.bst: No hyphenation pattern has been}%
\typeout{** loaded for the language `#1'. Using the pattern for}%
\typeout{** the default language instead.}%
\else
\language=\csname l@#1\endcsname
\fi
#2}}
\providecommand{\BIBdecl}{\relax}
\BIBdecl

\bibitem{burdickCharacterizationControlSelfmotions1989}
J.~Burdick and H.~Seraji, ``Characterization and control of self-motions in
  redundant manipulators,'' in \emph{Proceedings of the {{NASA Conference}} on
  {{Space Telerobotics}}, {{Volume}} 2}, 1989.

\bibitem{groomRealtimeFailuretolerantControl1999}
K.~Groom, A.~Maciejewski, and V.~Balakrishnan, ``Real-time failure-tolerant
  control of kinematically redundant manipulators,'' \emph{IEEE Transactions on
  Robotics and Automation}, vol.~15, no.~6, pp. 1109--1115, Dec. 1999.

\bibitem{almarkhiMaximizingSizeSelfMotion2019}
A.~A. Almarkhi and A.~A. Maciejewski, ``Maximizing the {{Size}} of
  {{Self-Motion Manifolds}} to {{Improve Robot Fault Tolerance}},'' \emph{IEEE
  Robotics and Automation Letters}, vol.~4, no.~3, pp. 2653--2660, Jul. 2019.

\bibitem{lewisFaultTolerantOperation1997}
C.~Lewis and A.~Maciejewski, ``Fault tolerant operation of kinematically
  redundant manipulators for locked joint failures,'' \emph{IEEE Transactions
  on Robotics and Automation}, vol.~13, no.~4, pp. 622--629, Aug. 1997.

\bibitem{xieMaximizingProbabilityTask2022}
B.~Xie and A.~A. Maciejewski, ``Maximizing the {{Probability}} of {{Task
  Completion}} for {{Redundant Robots Experiencing Locked Joint Failures}},''
  \emph{IEEE Transactions on Robotics}, vol.~38, no.~1, pp. 616--625, Feb.
  2022.

\bibitem{tatliciogluAdaptiveControlRedundant2009}
E.~Tatlicioglu, D.~Braganza, T.~C. Burg, and D.~M. Dawson, ``Adaptive control
  of redundant robot manipulators with sub-task objectives,'' \emph{Robotica},
  vol.~27, no.~6, pp. 873--881, Oct. 2009.

\bibitem{yaoMotionPlanningAlgorithms2010}
Y.~Yao, ``Motion {{Planning Algorithms}} of {{Redundant Manipulators Based}} on
  {{Self-motion Manifolds}},'' \emph{Chinese Journal of Mechanical
  Engineering}, vol.~23, no.~01, p.~80, 2010.

\bibitem{maaroofGeneralSubtaskController2012}
O.~W. Maaroof, E.~Gezgin, and M.~{\.I}. Can~Dede, ``General subtask controller
  for redundant robot manipulators,'' in \emph{2012 12th {{International
  Conference}} on {{Control}}, {{Automation}} and {{Systems}}}, Oct. 2012, pp.
  1352--1357.

\bibitem{pingStudyKinematicsOptimization2006}
Y.~Ping, S.~Hanxu, and J.~Qingxuan, ``Study on {{Kinematics Optimization}} of
  {{Redundant Manipulators}},'' in \emph{2006 {{IEEE Conference}} on
  {{Robotics}}, {{Automation}} and {{Mechatronics}}}, Jun. 2006, pp. 1--6.

\bibitem{maaroofPhysicalHumanRobotInteraction2016}
O.~W. Maaroof and M.~{\.I}.~C. Dede, ``Physical {{Human-Robot Interaction}}:
  {{Increasing Safety}} by {{Robot Arm}}'s {{Posture Optimization}},'' in
  \emph{{{ROMANSY}} 21 - {{Robot Design}}, {{Dynamics}} and {{Control}}},
  V.~{Parenti-Castelli} and W.~Schiehlen, Eds.\hskip 1em plus 0.5em minus
  0.4em\relax Cham: Springer International Publishing, 2016, vol. 569, pp.
  329--337.

\bibitem{zhangNovelDivisionBased2007}
C.~Zhang, H.~Sun, Q.~Jia, and L.~Hong, ``A {{Novel Division Based Self-Motion
  Algorithm}} for {{Avoiding Obstacles}} for {{Redundant Manipulators}},'' in
  \emph{2007 {{IEEE International Conference}} on {{Automation}} and
  {{Logistics}}}.\hskip 1em plus 0.5em minus 0.4em\relax Jinan, China: IEEE,
  Aug. 2007, pp. 852--857.

\bibitem{dasInverseKinematicAlgorithms1988}
H.~Das, J.-E. Slotine, and T.~Sheridan, ``Inverse kinematic algorithms for
  redundant systems,'' in \emph{Proceedings. 1988 {{IEEE International
  Conference}} on {{Robotics}} and {{Automation}}}.\hskip 1em plus 0.5em minus
  0.4em\relax Philadelphia, PA, USA: IEEE Comput. Soc. Press, 1988, pp. 43--48.

\bibitem{escandeHierarchicalQuadraticProgramming2013}
A.~Escande, N.~Mansard, and P.-B. Wieber, ``Hierarchical {{Quadratic
  Programming}}: {{Companion}} report,'' LJK, Research {{Report}}, 2013.

\bibitem{havilandManipulatorDifferentialKinematics2023}
J.~Haviland and P.~Corke, ``Manipulator {{Differential Kinematics}}: {{Part
  I}}: {{Kinematics}}, {{Velocity}}, and {{Applications}},'' \emph{IEEE
  Robotics \& Automation Magazine}, pp. 2--11, 2023.

\bibitem{peidroMethodBasedVanishing2018}
A.~Peidr{\'o}, {\'O}.~Reinoso, A.~Gil, J.~M. Mar{\'i}n, and L.~Pay{\'a}, ``A
  method based on the vanishing of self-motion manifolds to determine the
  collision-free workspace of redundant robots,'' \emph{Mechanism and Machine
  Theory}, vol. 128, pp. 84--109, Oct. 2018.

\bibitem{wuNovelMethodComputing2023}
T.~Wu, J.~Zhao, and B.~Xie, ``A novel method for computing self-motion
  manifolds,'' \emph{Mechanism and Machine Theory}, vol. 179, p. 105121, Jan.
  2023.

\bibitem{UFACTORYXArmUFACTORY}
``{{UFACTORY xArm}} 7 - {{UFACTORY Official Website}}.''

\bibitem{2020SciPy-NMeth}
P.~Virtanen, R.~Gommers, T.~E. Oliphant, M.~Haberland, T.~Reddy, D.~Cournapeau,
  E.~Burovski, P.~Peterson, W.~Weckesser, J.~Bright, S.~J. {van der Walt},
  M.~Brett, J.~Wilson, K.~J. Millman, N.~Mayorov, A.~R.~J. Nelson, E.~Jones,
  R.~Kern, E.~Larson, C.~J. Carey, {\.I}.~Polat, Y.~Feng, E.~W. Moore,
  J.~VanderPlas, D.~Laxalde, J.~Perktold, R.~Cimrman, I.~Henriksen, E.~A.
  Quintero, C.~R. Harris, A.~M. Archibald, A.~H. Ribeiro, F.~Pedregosa, P.~{van
  Mulbregt}, and {SciPy 1.0 Contributors}, ``{{SciPy}} 1.0: {{Fundamental}}
  algorithms for scientific computing in python,'' \emph{Nature Methods},
  vol.~17, pp. 261--272, 2020.

\bibitem{bohigasCompleteMethodWorkspace2012}
O.~Bohigas, M.~Manubens, and L.~Ros, ``A {{Complete Method}} for {{Workspace
  Boundary Determination}} on {{General Structure Manipulators}},'' \emph{IEEE
  Transactions on Robotics}, vol.~28, no.~5, pp. 993--1006, Oct. 2012.

\end{thebibliography}
